%% file: main.tex
\documentclass{article}

     \PassOptionsToPackage{numbers, compress}{natbib}

\usepackage[preprint]{neurips_data_2021}

\usepackage[utf8]{inputenc} 
\usepackage[T1]{fontenc}    
\usepackage{hyperref}       
\usepackage{url}            
\usepackage{booktabs}       
\usepackage{amsfonts}       
\usepackage{nicefrac}       
\usepackage{microtype}      
\usepackage{wrapfig}

\newcommand{\felix}[1]{\textit{\color{red}[Felix: ]}}
\newcommand{\thom}[1]{\textit{\color{red}[Thom: ]}}

\title{BEAMetrics: A Benchmark for Language Generation Evaluation Evaluation}

\author{%
  Thomas Scialom \\
  Sorbonne Universit\'e, CNRS, LIP6, F-75005\\
  reciTAL, Paris, France \\
  \texttt{thomas@recital.ai} \\
   \And
   Felix Hill \\
   DeepMind \\
   \texttt{felixhill@google.com} \\
}

\usepackage{graphicx}
\usepackage{comment}
\usepackage{multirow}
\usepackage{enumitem}
\usepackage{booktabs}
\usepackage[table]{xcolor}

\begin{document}

\maketitle

\begin{abstract}

Natural language processing (NLP) systems are increasingly trained to generate open-ended text rather than classifying between responses. This makes research on evaluation \emph{metrics} for generated language -- functions that score system output given the context and/or human reference responses -- of critical importance. However, different metrics have different strengths and biases, and reflect human intuitions better on some tasks than others. There is currently no simple, unified way to compare, analyse or evaluate metrics across a representative set of tasks. Here, we describe the Benchmark to Evaluate Automatic Metrics (BEAMetrics), a resource to make research into new metrics \emph{itself easier to evaluate}. BEAMetrics users can quickly compare existing and new metrics with human judgements across a diverse set of tasks, quality dimensions (fluency vs. coherence vs. informativeness etc), and languages. As generation experts might predict, BEAMetrics reveals stark task-dependent differences between existing metrics, and consistently poor performance on tasks with complex answer spaces or high reliance on general knowledge. While this analysis highlights a critical issue facing current research practice, BEAMetrics also contribute to its resolution by facilitating research into better metrics -- particularly those that can account for the complex interaction between context and general knowledge inherent to many modern NLP applications.\footnote{BEAMetrics is available under the MIT License: \url{https://github.com/ThomasScialom/BEAMetrics}}
\end{abstract}

\section{Introduction}

In the past, natural language generation (NLG)~\cite{reiter1997building} was important only for a subset Natural Language Processing (NLP) applications. Today, driven in part by the success of large autoregressive language models~\citep{brown2020language}, many NLP applications involve models that generate running text from open-ended vocabularies (see e.g.~\citep{chopra2016abstractive,cho2021unifying,tsimpoukelli2021multimodal}). Beyond classic NLG applications like machine translation and summarization, tasks that were previously framed as classification problems -- such as open-domain question answering -- are also now addressed by NLG systems. This trend is undoubtedly positive, since it points towards more general, flexible and expressive language technology. However, one side-effect is that NLP as a whole inherits the challenges inherent to NLG; in particular the issue of how to effectively evaluate systems that generate open-ended text. 

Evaluating NLG systems is notoriously difficult \citep{novikova2017we, pmlr-v119-scialom20a, sai2020survey}. The `ideal' approach (matching the setting of a deployed system) is to show system outputs to humans and have them rate their quality along different dimensions such as 'fluency` or `informativeness`. However, doing so is expensive, time-consuming and can be hard to replicate. Research therefore also relies critically on automatic evaluation metrics: functions that score system outputs given either human reference responses, the input context, or both. 

\begin{wrapfigure}{r}{0.5\textwidth}
    \centering \small
    \includegraphics[width=0.50\textwidth]{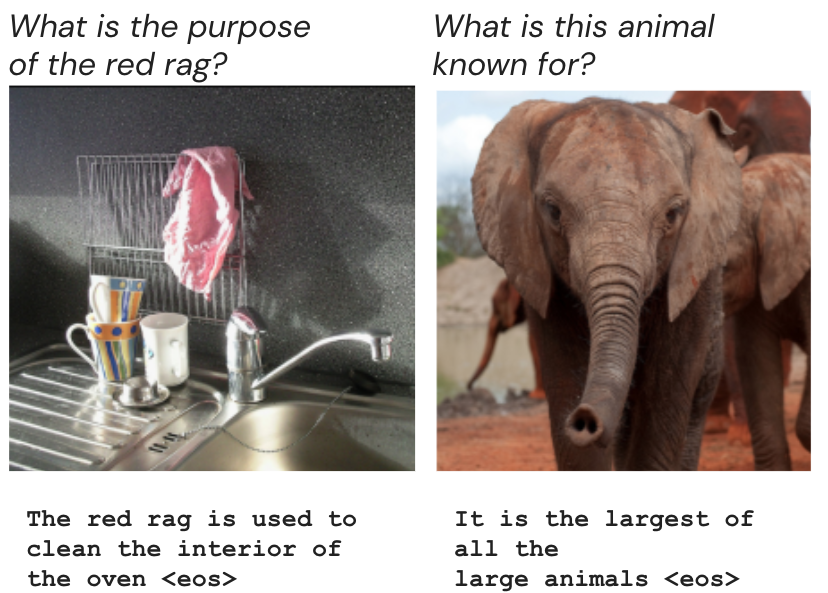}
    \caption{\textbf{The challenge faced by metrics evaluating modern NLG systems} Example output from a system built with CLIP~\citep{radford2021learning} and GPT2~\citep{radford2019language} on the OK-VQA~\citep{marino2019ok} task. The next generation of metrics must find a way to reasonably assign credit to such responses given the available context and/or reference human responses.}
    \label{fig:okvqa_examples}

\end{wrapfigure}

Automatic metrics are generally quick to apply and consistent 
making them essential for enabling fast iteration when designing systems. Despite their obvious utility, however, there is no known metric (or set of metrics) that adequately reflects human intuitions across the growing spectrum of NLG-relevant tasks~\citep{sai2020survey}. Moreoever, as illustrated in Figure~\ref{fig:okvqa_examples}, many of the new application areas for NLG models pose the greatest challenges for evaluation. Given that NLG research depends critically on metrics, and NLP as a whole relies increasingly on NLG, the need for new approaches to evaluation metrics has never been greater.\footnote{This view is certainly shared, for instance, by \citet{ruder2021benchmarking}, who argues that \emph{we need to rethink how we design our benchmarks and evaluate our models so that they can still serve as useful indicators of progress.}}


With this motivation, we propose \emph{BEAMetrics}, the Benchmark to Evaluate Automatic Metrics.
BEAMetrics is a resource for \emph{evaluation evaluation} - a simple and easy way to fairly compare and evaluate evaluation metrics. With the code provided, users of BEAMetrics can quickly and easily compare the predictions of new NLG metrics against a fixed set of human judgements across different quality dimensions, NLG tasks and languages. It therefore removes any need for the developers of new evaluation metrics to conduct their own human evaluations, and in doing so provides a consistent basis for detailed analysis and replicable comparisons. As such, we hope that BEAMetrics will motivate a concerted focus among researchers to consider and improve evaluation metrics, in much the same way that the ImageNet Challenge \cite{deng2009imagenet} and GLUE \cite{wang2018glue} stimulated progress in image and language classification respectively. 

\section{Related Work}
\subsection{Benchmarks}
Benchmarks have always played a key role in both guiding and measuring the progress of AI research. The ImageNet challenge ~\citep{russakovsky2015imagenet}, for instance, played an important role in revealing the potential of Deep Learning \cite{krizhevsky2012imagenet}. Many benchmarks today assess the generality of a system, by aggregating suites of related tasks. A notable example of this is the recent Abstraction and Reasoning Corpus (ARC) \citep{chollet2019measure}, psychometric intelligence test composed 600 unique tasks, to measure the level of general intelligence in artificial systems. 

Language is a highly general and multi-faceted domain, and benchmarks are playing an increasingly important role in NLP: the General Language Understanding Evaluation benchmark (GLUE) 
contributing to  highlight the potential of self-supervised learning for language problems \cite{radford2018improving, devlin2018bert}. GLUE encompasses 9 different Language \emph{classification} tasks including sentiment analysis \cite{socher2013recursive}, sentence similarity \citep{dolan2005automatically, cer2017semeval}, Natural Language Inference \citep{dagan2005pascal, williams2017broad}, Question Answering \citep{rajpurkar2016squad}, and coreference \citep{levesque2012winograd}. While state-of-the-art approaches have now achieve a human-level performance on GLUE \citep{liu2019roberta, lewis2019bart, raffel2019exploring}, \citet{wang2019superglue} proposed a more challenging successor, SuperGLUE.

Most recently,  \citet{gehrmann-etal-2021-gem} proposed GEM, a benchmark specifically for tasks requiring Natural Language \emph{Generation}. GEM encompasses 4 tasks through 11 datasets: Summarization \citep{scialom-etal-2020-mlsum, narayan2018don, ladhak-wiki-2020}, Structure To Text \citep{gardent2017creating, lin-etal-2020-commongen, e2e_cleaned, radev2020dart, cs_restaurants, parikh2020totto}, Dialogue \citep{rastogi2019towards}, and Simplification \citep{jiang-etal-2020-neural, Xu-EtAl:2016:TACL, alva-manchego-etal-2020-asset}. 
As noted in the introduction, evaluation of generated language relies not only on tasks, but also on automatic metrics (or humans). The developers of GEM provide trusted implementations of known metrics such as BLEU, ROUGE, BERTscore etc\footnote{https://github.com/GEM-benchmark/GEM-metrics}. However, because there is currently no way to assess the trustfulness of metrics, GEM also encourages human evaluation during the workshops,~\footnote{https://gem-benchmark.com/workshop}.

\subsection{Evaluating Metrics}
\label{subsec:evaluating_the_metrics}
While human evaluation will always be valuable, practicalities mean that effective automatic metrics can play a critical role in guiding and enabling research on NLG problems. The limitations of current automatic metrics, however, are well known. \citet{callison2006re} showed that a BLEU improvement is neither necessary nor sufficient for achieving an actual improvement. \citet{louis2013automatically} showed similar findings regarding other metrics like ROUGE. \citet{choshen2018inherent} demonstrated the inherent presence of biases in reference-based evaluation.

\citet{novikova2017we} make an impassioned appeal for `why we need new evaluation metrics for NLG'. They investigated a wide range of metrics, and reported that human judgement is only weakly reflected. Importantly, they also showed that metric performance is very often task, and even dataset dependent. This finding indicates the importance of evaluating and understanding the performance of metrics on \emph{multiple tasks}.

While metrics may be valuable for evaluating any NLG task, many of the best known metrics, like BLEU score, were primarily designed for evaluating Machine Translation (MT) systems. In a large and comprehensive analysis of metrics in MT, however, \citep{kocmi2021ship} recommends deprecating BLEU as the MT evaluation standard. They suggest using more recent metrics such as BERTScore \citep{zhang2019bertscore} or COMET \citep{rei2020comet}, that are shown to reflect human judgement significantly better. 

Given the impressive performance on MT for BERTScore, its adoption on other tasks has now started, for instance in Summarization \citep{li2019deep,  leblond2021machine, paraschiv2020upb, ju2020scisummpip, zhang2021emailsum}. Unfortunately, BERTScore actually performs weakly on Summarization \citep{fabbri2021summeval, Bhandari-2020-reevaluating}. This pattern reflects a general tendency for researchers to apply a metric designed or proven for one NLG task to some other task, in apparent ignorance of its limited scope. This in turn emphasises the need for an easy and replicable way to quantify the performance of new or existing metrics across a representative sample of NLG tasks. Such a resource would also encourage the research community to develop metrics that generalize beyond only one task.

\section{The BEAMetrics benchmark}
\label{sec:beametrics_benchmark}

\begin{table}[]
\tiny \centering
\begin{tabular}{|l|ccccc|}
\toprule
\textbf{Dataset} & \textbf{Abbreviation} & \textbf{Task}                                                  & \textbf{Languages}                                                  & \textbf{Dimensions}  & \textbf{Nb Ex}  \\
\midrule
WebNLG           & Web                   & Data2text                                                      & EN                                                                    & Cor, Flu, Gra       & 2,007\\
\midrule
Asset-Eval       & Asv                   & Simplification                                                 & EN                                                                        & Cor, Flu, Sim    & 162   \\
\midrule
MUSS             & MUS                   & Simplification                                                 & EN, FR, ES                                                                & Cor, Flu  & 150          \\
\midrule
WMT2019          & WMT                   & \begin{tabular}[c]{@{}c@{}}Machine \\ Translation\end{tabular} & \begin{tabular}[c]{@{}c@{}}EN, DE, FI, GU, \\ KK, LT, RU, ZH\end{tabular} & Cor  & 480,332               \\
\midrule
Pascal50s        & Pas                   & Captioning                                                     & EN                                                                        & Cor & 4,000                \\
\midrule
Flickr8k         & Fli                   & Captioning                                                     & EN                                                                        & Cor  & 5,664               \\
\midrule
RealSum          & Rea                   & Summarization                                                  & EN                                                                        & Cor  & 2,500               \\
\midrule
SummEval         & SumE                  & Summarization                                                  & EN                                                                        & Cor, Rel, Coh, Flu &  1,600 \\
\midrule

MultiSummEval    & mSu                   & Summarization                                                  & \begin{tabular}[c]{@{}c@{}}EN, DE, ES, FR, \\ RU, TR, ZH, ID\end{tabular} & Cor, Rel  & 2,160          \\
\midrule
Efficient QA     & OpQA                  & QA                                                             & EN                                                                        & Cor, Obv     & 9,000       \\
\midrule
OkVQA            & OkVQA                 & VQA                                                            & EN                                                                        & Cor, Obv, Pos & 300   \\  
\bottomrule
\end{tabular}

\label{tab:datasets_in_beametrics}
\vspace{0.1cm}
\caption{Datasets included in the BEAMetrics benchmark, and their characteristics. They are detailed on Section \ref{subsec:selected_datasets}. The dimensions are defined in \ref{subsec:results_display}; Cor stands for Correctness, Flu: Fluency, Gra: Grammar, Sim: Simplicity, Rel: Relevance, Coh: Coherence, Obv: Obviousness, and Pos: Possibility.}
\vspace{-0.5cm}
\end{table}

\subsection{Design principles}
\label{sec:beametrics_general_guidelines}

In constructing a benchmark for metrics, we set out various desiderata and priorities based on what has worked well in other benchmarks in the literature. In particular, the benchmark should reflect the metric generalisation across three main axes: tasks, languages, and evaluated dimensions:  

\paragraph{Multilingual} NLP research has mostly focused on English while 7000+ languages are spoken around the world.\footnote{\url{https://ruder.io/nlp-beyond-english/}}. Token level metrics like ROUGE or BERTScore are not suited to all the language morphologies   \citep{lee-etal-2020-reference}. We want \emph{BEAMetrics} to measure the multilingual ability of a metric.

\paragraph{Multitask} As noted above, the panorama or NLG-relevant tasks in NLP is rapidly expanding. The performance of metrics should be known on a representative sample of these tasks. In this aspect, BEAMetrics draws on the impact of general multitasks benchmarks like GLUE.

\paragraph{Evaluated Dimensions} In many cases, the quality of textual output can (and should) be evaluated along different dimensions, as this better capturing the multi-faceted function and nature of language. Overall, BEAMetrics contains evaluations for 9 such dimensions: adequacy, correctness, fluency, simplicity, grammar, relevance, coherence, possibility, and obviousness. Each of these dimensions are defined in the data-card of the corresponding dataset (see Section \ref{subsec:datastatement} about the data-cards).

Beyond these three axes, we also aimed for high diversity among the collected data:
\begin{itemize}[noitemsep, leftmargin=*]
    \item different dataset covering various domains, and with different characteristics, e.g. long Vs short references; 
    \item diversity in the systems evaluated: state-of-the-art systems as well as older; extractive and abstractive models;
    \item different evaluation protocols, based on different scales (likert, ranking). 
\end{itemize}

This diversity should add the robustness of conclusions drawn from BEAMetrics, and help to stimulate the development of increasingly general metrics - potentially sharing knowledge across 'tasks'.


Finally, the datasets we selected in BEAMetrics have all been peer reviewed and publicly released. 

Overall, across its 11 datasets, BEAMetrics is multi[task, lingual and dimensional], as we show in Table~\ref{tab:datasets_in_beametrics}.
For each dataset, we provide a complete data statement, as presented in the next Section. 


\subsection{Data Statement}
\label{subsec:datastatement}

To mitigate system bias, improve reproducibility and enable better science, \citep{bender-friedman-2018-data} and \citep{Gebru2018datasheets} recommend implementing a Data Statement. It consists of a standardized process of documenting the datasets.

We therefore propose a template card that contains the important information about a human evaluation dataset: the evaluation set(s) used to generate the texts, the language(s), the task, the number of human references available for each examples, the evaluated dimensions and their exact definition in the original paper, the annotation protocol and the rating scale, information about the annotators, any additional comments, and finally the citation of the paper.

For each dataset that composed BEAMetrics, we make available the template card completed.\footnote{\url{https://github.com/ThomasScialom/BEAMetrics/tree/main/data/datacards}}
Note that a data card is directly integrated in the BEAMetrics code: it is generated by filling our proposed template with the information that are provided in the configuration file of the dataset. This makes the process both standardized and mandatory, when a a new dataset is added to the suite.

\subsection{Selected Datasets}
\label{subsec:selected_datasets}

\textbf{WMT2019 (WMT)} We consider 7 language pairs from the WMT19 metrics shared task \citep{ma-etal-2019-results}: de-en, fi-en, gu-en, kk-en, lt-en, ru-en, and zh-en. We rely on the DaRR corpus corresponding to an evaluation via Direct Assessment from 1 to 100, followed by Relative Ranking over 480,332 outputs.

\textbf{SummEval (SumE)} \citep{fabbri2021summeval} is one the largest studies in Summarization. It includes the evaluation of 1,600 generated summaries on the CNN/DM corpus \citep{nallapati2016abstractive} from 16 models on 100 source documents ($16 * 100=1600$). The annotators rated the summaries on four dimensions: Fluency, Correctness, Coherence and Relevance. A notable aspect of this study is the use of expert annotators, which is important given the difficulty of evaluating summaries, even for humans. 

\textbf{SumEval-multi (mSu)} \citep{koto2021evaluating} is a recent multilingual extension of SummEval. It contains the evaluation of 2,160 summaries from four multilingual datasets including a total of 8 languages: French, Spanish, German, Russian, and Turkish from MLSUM \citep{scialom-etal-2020-mlsum}, Indonesian from Liputan6 \citep{koto-etal-2020-liputan6}, English from CNN/DM \citep{chopra2016abstractive}, and Chinese from LCSTS \citep{hu-etal-2015-lcsts}.
Two dimensions are evaluated: the precision and the recall for the outputs of 2 abstractive systems: i) Pointer Generator \citep{see2017get} a model with copy mechanism and not pretraining, and ii) BERT-gen \citep{scialom2020bert}.

\textbf{REALSumm (Rea)} \citep{Bhandari-2020-reevaluating} is an evaluation on 2,500 system generated summaries, that includes outputs for 14 abstractive and 11 extractive models. The annotation protocol is based on the lightweight-pyramid \citep{shapira-etal-2019-crowdsourcing}, a cost effective adaption of the pyramid \citep{nenkova-passonneau-2004-evaluating} method that allows to use Mechanical Turk. In lightweight-pyramid, the time consuming merge of duplicate Semantic Content Units (SCU) from different reference summaries is replaced by a simple SCU sampling. 

\textbf{Asset-eval (Asv))} \cite{alva2020asset}
is a corpus of a total of 9,000 ratings that was released along with the ASSET corpus. Mechanical Turkers rated on a Likert Scale the Fluency, Correctness, and Simplicity of the system-generated simplifications.

\textbf{MUSS-eval (MUS)}  \citep{martin2020muss} is a human evaluation of 150 text simplification outputs in three languages (English, French, and Spanish). The annotators were volunteers that rated each sample over three dimensions: Simplicity, Adequacy, and Fluency.

\textbf{WebNLG-eval (Web):} \cite{shimorina2018webnlg}
consists of a set of 2,000 English descriptions of structured tables generated by 10 different systems and annotated on their Fluency, Correctness and Grammar. This corpus was provided along with the WebNLG Data-To-Text generation corpus, which consists of tables and corresponding descriptions in 16 DBPedia categories (e.g., Airport, Astronaut, etc.). 

\textbf{Flickr8k (Fli):} \cite{hodosh2013framing} is an image captioning evaluation, where three human-expert annotators rated the relevance of 5,822 generated captions regarding their source image, from 1 to 4.
            
\textbf{PASCAL50S (Pas):} \cite{vedantam2015cider} is an image captioning evaluation over 4k examples. It evaluation is a comparison between two generated captions: the annotators were asked to judge which caption is more appropriated given 50 references of the given image. 

\textbf{NeurIPS Question Answering (OpQA)} \cite{min2021neurips} corresponds to the human evaluation conducted for the Efficient QA track at the NeurIPS 2020 competition.\footnote{\url{https://efficientqa.github.io/}} The Top 5 QA systems were manually evaluated over 1,800 questions. The raters marked each of the 9,000 generated answer ($1,800*5$) as either `definitely correct', `plausibly correct', or `definitely incorrect'. 
We therefore consider the two following dimensions: i) \emph{Correctness}: True if the answer is definitely or plausibly correct, and False if it is incorrect. ii) \emph{Obviousness}: True if the answer is definitely correct, and False otherwise.

\textbf{OKVQA-Eval (OkVQA)} OKVQA \citep{marino2019ok} is a VQA dataset with questions about images that require Outside Knowledge to be integrated with the information in the image, see for instance two examples in Figure \ref{fig:okvqa_examples}. We chose to conduct a human evaluation on OKVQA responses and to include them in BEAMetrics, since this task emphasises an important challenge that metrics of the future may need to address. The information required to resolve an OKVQA question is not entirely contained within the context image, so that answering questions requires 'outside knowledge'. Consequently, a complete evaluation of model output must also require outside knowledge (See \ref{subsec:discussion} for discussion of this point). 

To get reasonable model responses on OKVQA for humans to assess, we generated answers using an NLG system that combines two well-known large-scale data-intensive models: CLIP \citep{radford2021learning} and GPT-2 \citep{radford2019language}. The model integrates CLIP ouput into the GPT-2 input space via a non-convex combination of word embedddings.\footnote{The model was developed by Jamie Kiros~\citep{kiros}, and is available at \url{https://colab.research.google.com/drive/1fokumWeasHTo0KXpfeZ6Z0OgLOQ2SUso?usp=sharing}}.  

The annotators were given an image-question pair, and the generated answer, and were asked to evaluate three different dimensions:
\begin{enumerate}[noitemsep, leftmargin=*]
    \item Correctness: Is the answer definitely factually correct (use Google if necessary)?
    \item Possibility: In your opinion, is the answer possible in some possible situation?  
    \item Obviousness: If the image was shown to 100 people and the question was asked, how many people do you think would give the answer? This question  were inspired by the TV Show Family Fortune~\footnote{\url{https://en.wikipedia.org/wiki/Family_Fortunes}} and can be interpreted as representing the mass probability among the possible answers. 
\end{enumerate}
In term of inter-rater agreement, we obtain a Krippendorff alpha \citep{krippendorff2011computing} of 0.67 for Correctness, 0.74 for Obviousness and 0.48 for Possibility. The significantly higher alpha for Obviousness than Possibility indicates that annotators agree more on the mass probability among the possible answers allocated by other humans, rather than the Possibility of the given answer.

Additional details are given in the Appendix, including among others the full annotation protocol, a screenshot of the annotation tool, the average time to annotate an answer.

\section{Metrics}

Numerous metrics have been proposed to evaluate NLG systems. Several surveys have made interesting taxonomies \citep{sai2020survey, kocmi2021ship} to regroup metrics regarding some fundamental distinctions, such as being parametric (i.e. supervised) Vs rule-based, acting at the sequence level Vs at a token level, or the token representation being neural based Vs n-gram based. 

In this paper, our focus is to measure the ability of the metrics. Therefore, beyond there internal characteristics, we group the metrics regarding only the inputs they used to evaluate a text: 
\begin{enumerate}[noitemsep, leftmargin=*]
    \item Reference-based: metrics that leverage the reference(s) to evaluate the text. The source can also be used additionally;
    \item Reference-less: metrics that use only the source, without requiring any reference. Note that this setup corresponds to the namely Quality Estimation (QE) in MT.
\end{enumerate}


As baselines, we consider n-gram statistics, standard metrics like BLEU, ROUGE, or METEOR, and recent neural metrics like BERTScore, Nubia and BLEURT.

\textbf{BLEU} \citep{papineni2002bleu} measures the overlap of n-grams between the evaluated text and its reference(s).

\textbf{ROUGE} \citep{lin2004rouge} stands for Recall-Oriented Understudy for Gisting Evaluation. Similarly to BLEU, it is based on the count of overlapping n-grams, but it recall oriented.

\textbf{METEOR} \citep{lavie2007meteor} was proposed to fix some of the problems in BLEU. While BLEU seeks correlation at the corpus level, METEOR provide scores at the sentence level. 

\textbf{BERTScore} \citep{zhang2019bertscore} leverages the contextualised representation of BERT to compute the similarity between the tokens.

\textbf{BLEURT} \citep{sellam2020bleurt} is a learned evaluation metric based on BERT, then fine-tuned on WMT human annotations to emulate the annotators.

\textbf{Nubia} \citep{kane2020nubia} stands for NeUral Based Interchangeability Assessor. It is composed of three modules: first, a module that obtains different neural representations of the evaluated text, then an aggregator of the representations, and finally a calibration module.

\textbf{GPT2 Perplexity} \citep{radford2019language} We report the perplexity of the evaluated text using GPT2.   

\textbf{Statistics} We also report the correlations of simple heuristics: 
\begin{itemize}[noitemsep, leftmargin=*]
     \item \textbf{Length:} The number of n-grams in the evaluated text;
    \item \textbf{Repetitions:} the number of n-grams that are repeated multiple times, normalised by the length;
    \item \textbf{Abstractness:}  the number of n-grams not present in the source text, normalised by the length.
\end{itemize}

All the metrics are integrated in the BEAMetrics framework using either the official implementation (e.g. for Nubia), a trustable version (e.g. SacreBLEU \citep{post-2018-call}, the HuggingFace implementation \citep{wolf2019huggingface}.

\section{Results and Discussion}
\label{sec:results_discussion}

\input{main_table_correctness_pearson}

\subsection{Results Display}
\label{subsec:results_display}
As discussed in Section \ref{sec:beametrics_general_guidelines}, several different dimensions are represented in BEAMetrics. 
Averaging all the results into a single number can limit the interest of the benchmark, and could be an incentive for “leaderboard chasing” approaches.
On the other hand, the large number of datasets and evaluated dimensions make the final results difficult to interpret at a glance, if not clearly structured. 
For this reason, we propose to group the results into two distinct tables: 
First, \textbf{Correctness:} the correctness of the predication given the context. This is arguably the principal dimension: is of of the utmost importance for a text to be factually consistent with its context. This also explains why each evaluation set described in Section \ref{subsec:selected_datasets} has included this dimension in their evaluation.
In the second Table, we regroup the other dimensions: 
\begin{itemize}[noitemsep, leftmargin=*]
    \item Fluency: The quality of individual sentences.
    \item Coherence: The text should be well-structured and well-organized, not just be a heap of related information.
    \item Simplicity: In Simplification, is the text easy to read, and composed of simple words?
    \item Relevance: In Summarization, the selection of important content from the source.
    \item Possibility: In QA, is the answer possible?
    \item Obviousness: In QA, how expected is the answer, in the context?
\end{itemize}

We can consider the correlations among the different tasks for Correctness, as consistent all together, and hence also report the average score.
Conversely, we don't report any average for the second Table, as it contains correlations over unrelated dimensions. 

Finally, in each tables we will report the correlations in three separate blocs, that correspond to three different scenarios:
i) The top bloc contains reference-based metrics that have access to \emph{all} the references available in the corpus. It indicates the metric potential given a large number of references.
ii) The middle bloc, contains reference-based metrics that have access to \emph{only one} reference. It indicates the metric performance in a more likely scenario, given that most of modern datasets provide only a single reference. 
iii) At the bottom bloc, the reference-less metrics results. Those metrics are more challenging, but enable to evaluate a text without requiring a gold-reference.

\input{main_table_other_pearson}

\subsection{Results}

We report the Pearson Coefficients for the Correctness and the Other dimensions, respectively in Tables \ref{tab:main_res_correctness} and \ref{tab:main_res_other}. Note that we also report the Kendall Tau correlations and the p-values in the Appendix. 

As expected, we observe that BLEU performs worst than the neural metrics (i.e. BERTScore, Nubia and BLEURT). More surprising, the other n-gram based metrics, namely ROUGE and METEOR are actually competitive with neural metrics: on Correctness, BLEU obtains in average the worst performance (12.0), BLEURT performs the best (14.3), but only slightly above ROUGE1 and METEOR (14.2).
We hypothesise that the poor performance of BLEU can be explained by its overused trough time, a conclusion in line with a recent study \citep{kocmi2021ship}. 
This result also gives a new light and perspective about the performances of recent neural metrics compared to more standard ones.

An other unexpected result is the strong performance of BLEURT compared to the other neural metrics. BLEURT is a supervised metric trained to predict the annotator score of a given segments in MT. While we can expect BLEURT to compare favorably on MT, its relatively good performance beyond MT is more surprising. We note that few works outside MT have reported BLEURT, as opposed to BERTScore. It will be worth exploring the reasons behind this good performance in future works.

Finally, the average correlations are relatively low, which emphasize the need to develop better metrics. We hope that BEAMetrics will contribute in this direction, by providing off-the-shelf tools for researchers to evaluate evaluation metrics.

\subsection{The Future of Metrics}

\label{subsec:discussion}

\paragraph{Question answering and NLG} To the best of our knowledge, ours is the first treatment to consider in depth the evaluation challenges posed by systems that can answer questions (QA) with open-ended running text. Indeed, QA is not a domain included in GEM, so a fortiori has not been a major concern for those interested in the evaluation of generated language. This is in part because QA was in the past normally framed as a classification task e.g. in SQuAD \citep{rajpurkar2016squad}. Such a constraint does not yield particularly flexible or expressive systems, however, and the advent of better generative language models has recently placed both open-domain \citep{roberts2020much,prager2006open} and visual~\citep{cho2021unifying,tsimpoukelli2021multimodal} QA tasks squarely within the scope of NLG.

As illustrated in Figure~\ref{fig:okvqa_examples}, question-answering in its most general form poses arguably the greatest evaluation challenge of any NLG domain. For some questions, there may be a single correct answer (a factoid), in which case a metric must simply account for superficial variation in how that answer may be expressed. However, for many other questions, there can be a complex space of possible answers, characterised by gradual (rather than discrete) variation of along quality dimensions \emph{and} multiple distinct answer `modes' (semantically distinct answers that are equally correct). This complexity points to the need for knowledge-driven metrics that are themselves complex (potentially non-linear) functions learned themselves from large and diverse human data. Perhaps unsurprisingly, then, on the QA and VQA BEAMetrics tasks we observe very low correspondence with human intuitions across all current automatic metrics. Among those, even the most expressive learned metric, BLEURT, was only trained on 
a single, very different, task (MT). By including two question-answering tasks in BEAMetrics, we emphasize the importance of developing smarter, more expressive metrics to support research on truly open-ended, flexible NLG systems. 

\begin{wrapfigure}{r}{0.5\textwidth}
    \centering \small
    \includegraphics[width=0.4\textwidth]{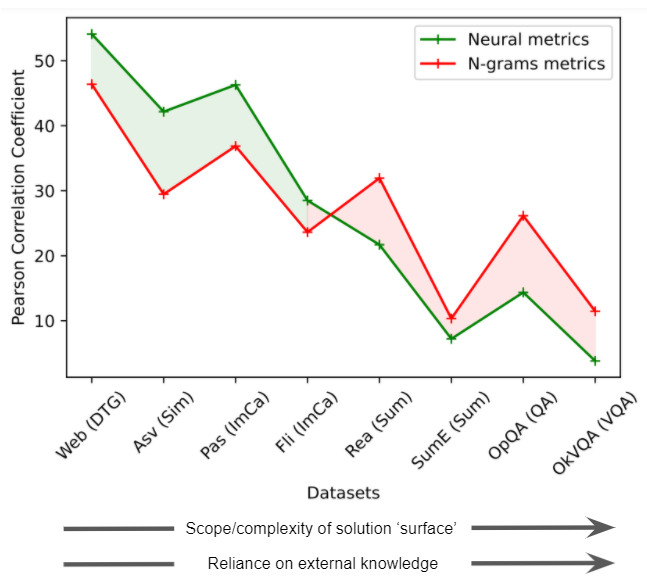}
    \caption{\textbf{The performance of metrics drops substantially for tasks with a more complex answer space and/or that require greater general knowledge}. Pearson correlation with human ratings, on the Correctness dimension for the tasks in English. Green line: an average of  n-gram based metrics (i.e. ROUGE, BLEU, METEOR); Red line, average of neural based metrics (i.e. BERTScore, BLEURT, Nubia).}
    \label{fig:information_assymetry}
\end{wrapfigure}

\paragraph{A problematic trend} This consideration of what makes QA a difficult setting to evaluate NLG systems also points to a general axis of variation among the tasks included in BEAMetrics. For classical NLG settings like MT, the space of responses for a given stimulus is relatively unimodal. In addition, to evaluate a translation, one may require comparatively little general knowledge beyond what information is contained in the stimulus and reference responses. A task like abstractive summarization increases the difficulty along both of these axes. For summarization, there may be two equally good summaries that differ somewhat in meaning (i.e. the answer space is \emph{less} unimodal). Further, good abstractive summarization arguably requires a degree more general knowledge than in the case for MT. While image captioning may require somewhat less general knowledge, the answer space is clearly not unimodal - various semantically distinct captions can be appropriate for a given image. Question answering can often exhibit the highest degree of difficulty along both of these axes. As shown in Figure~\ref{fig:information_assymetry}, the performance of current evaluation metrics (both n-gram based and neural) degrades quite consistently as one moves along these axes. Importantly, this phenomenon seems amplified for neural metrics, which actually perform worse than n-gram ones on Summarization, Open-ended QA and OkVQA. 

This analysis suggests that developing good metrics for MT -- where most research on metrics to date has focused -- may be a categorically different challenge than developing those for other types of NLG systems. Research on metrics for QA must be expressive enough to model a complex distribution of potential answers, and may have to focus to a greater extent than current approaches on both the integration of general knowledge and the consideration of both context. Of course, it is possible that the task of perfectly evaluating an open-ended NLG system, is almost as challenging as performing the task itself. A metric does not have to be perfect to be valuable, however; but it does need to reach a certain level of quality to provide a useful signal for guiding research.

\subsection{Limitations and Maintenance Strategy}


While systems performance improve, and the quality of their outputs increases, it becomes more challenging for metrics to reflect human preference \citep{scialom2021rethinking}.
Therefore, it is important to keep evaluating the metrics not only on outdated systems, but also on recent state-of-the-art systems. 
Therefore, we plan to progressively integrate new human evaluations in BEAMetrics, as the filed will evolve.




In this work we have presented BEAMetrics, a Benchmark to Evaluate Automatic Metrics. We hope it will serve the community, both as vehicle for better understanding the strengths and limitations of current metrics across a broad spectrum of tasks, as well as a tool to drive research into future metrics that better address the challenges of evaluating flexible generative models of language. BEAMetrics aims to place these evaluation challenges at the forefront of the minds of more researchers, encouraging a comparable level of focus on creative solutions for evaluation, as there is currently on developing new systems. After all, before we can make things better, we need robust, reliable, consistent and efficient ways of quantifying what better actually means.


\bibliography{neurips}
\bibliographystyle{icml2020}

\section*{Checklist}

The checklist follows the references.  Please
read the checklist guidelines carefully for information on how to answer these
questions.  For each question, change the default \answerTODO{} to \answerYes{},
\answerNo{}, or \answerNA{}.  You are strongly encouraged to include a {\bf
justification to your answer}, either by referencing the appropriate section of
your paper or providing a brief inline description. 

\begin{enumerate}

\item For all authors...
\begin{enumerate}
  \item Do the main claims made in the abstract and introduction accurately reflect the paper's contributions and scope?
    \answerYes{}
  \item Did you describe the limitations of your work?
    \answerYes{}, in the discussion, see Section \ref{subsec:discussion}.
  \item Did you discuss any potential negative societal impacts of your work?
    \answerNA{} 
  \item Have you read the ethics review guidelines and ensured that your paper conforms to them?
    \answerYes{}
\end{enumerate}

\item If you ran experiments (e.g. for benchmarks)...
\begin{enumerate}
  \item Did you include the code, data, and instructions needed to reproduce the main experimental results (either in the supplemental material or as a URL)?
    \answerYes{}, both in the supplementary material and an URL on the first page.
    \item Did you report error bars (e.g., with respect to the random seed after running experiments multiple times)?
    We report the p-values for the significance for all our experiments in the Appendix.
\end{enumerate}

\item If you are using existing assets (e.g., code, data, models) or curating/releasing new assets...
\begin{enumerate}
  \item If your work uses existing assets, did you cite the creators?
    \answerYes{}, all the dataset creators are cited, and a specific data card is also release. 
  \item Did you include any new assets either in the supplemental material or as a URL?
    \answerYes{}
  \item Did you discuss whether and how consent was obtained from people whose data you're using/curating?
    \answerYes{}, we restricted the  selection to publicly available data
\end{enumerate}

\item If you used crowdsourcing or conducted research with human subjects...
\begin{enumerate}
  \item Did you include the full text of instructions given to participants and screenshots, if applicable?
    \answerYes{} in the supplementary material.
  \item Did you include the estimated hourly wage paid to participants and the total amount spent on participant compensation?
    \answerYes{}, in the supplementary material as well.
\end{enumerate}

\end{enumerate}


\appendix

\section{Appendix}

\subsection{Example of a Data Card}

All the Data Cards are \href{https://github.com/ThomasScialom/BEAMetrics/blob/main/data/datacards}{\color{blue}publicly available}, see for instance \href{https://github.com/ThomasScialom/BEAMetrics/blob/main/data/datacards/SummmEval.md}{\color{blue}the SummEval card}.

\subsection{OkVQA annotation}

\begin{figure}
    \centering 
    \includegraphics[width=0.95\textwidth]{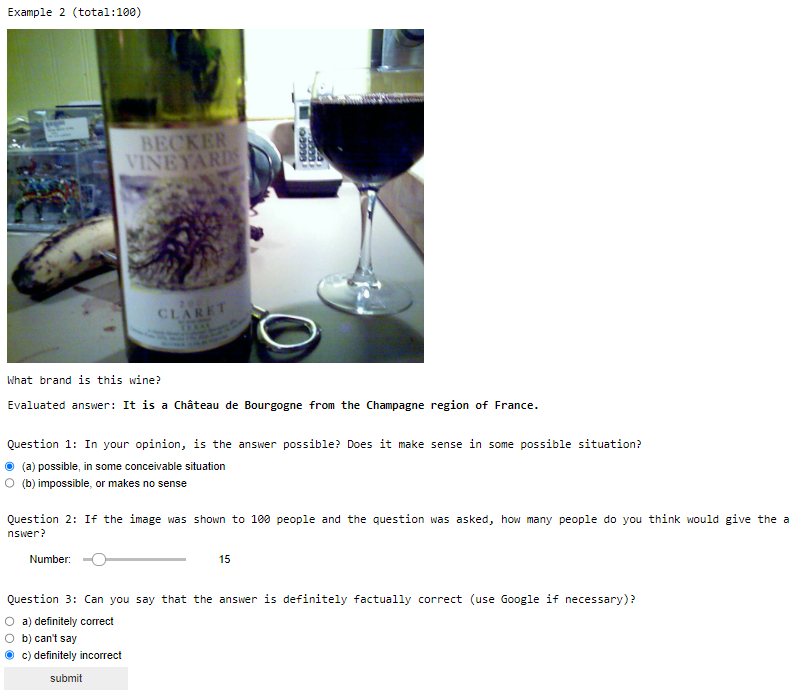}
    \vspace{-5pt}
    \caption{Print Screen of the annotation tool used for our human evaluation on No-VQA.}
    \label{fig:human_annot_tool}
\end{figure}

Three professional English Speakers were hired to evaluate the 300 answers.
They spend respectively 2h00min, 1h35min and 1h50min for the evaluation, and were compensated with vouchers.
They followed the protocol reported in Table \ref{tab:protocol_hum_annotation}
\\
\begin{table}[h]
    \centering
    \begin{tabular}{|p{8cm}|}
\tiny
\begin{verbatim}

Evaluation Protocol:

You are shown an image, a question and its answer. 
Your task is to evaluate the answer regarding three specific 
dimensions.

First, look at the image, read the question and then the 
answer.

Then, answer the following questions using the rating scale:

Question 1: In your opinion, is the answer possible? 
Does it make sense in some possible situation?
    (a) possible, in some conceivable situation
    (b) impossible, or makes no sense

Question 2: If the image was shown to 100 people and the 
question was asked, how many people do you think would give 
the answer?
    (write a number between 1 and 100)

Question 3: Can you say that the answer is definitely 
factually correct (use Google if necessary)?
    a) definitely correct
    b) can't say (matter of opinion, depends on the 
    situation, the question does not have a 'right' answer)    
    c) definitely incorrect
\end{verbatim}
\end{tabular}
    \caption{Human annotation protocol of the OkVQA dataset.}
    \label{tab:protocol_hum_annotation}
\end{table}

The protocol was followed by three examples, before the evaluation to start. We report in Figure \ref{fig:human_annot_tool} a print screen of the interface.

\subsection{Kendall Tau Correlations \& Confidence Scores}

In this Section, we report the main results similar to Tables \ref{tab:main_res_correctness} and \ref{tab:main_res_other} in Section \ref{sec:results_discussion}, but this time using Kendall Tau, instead of Pearson coefficient. Moreover, we also report the p-values for all the tables including the two based on Pearson, in the main paper.

\input{supp_material/main_table_correctness_kendall}

\input{supp_material/main_table_other_kendall}

\input{supp_material/main_table_correctness_kendall_confidence}

\input{supp_material/main_table_other_kendall_confidence}

\input{supp_material/main_table_correctness_pearson_confidence}

\input{supp_material/main_table_other_pearson_confidence}

\end{document}

%% file: main_table_correctness_pearson.tex
\begin{table}[h]
\tiny

\begin{tabular}{llcccccccccccc}
\toprule
{} & {Dataset:} & {WMT} & {Web} & {Asv} & {MUS} & {Pas} & {Fli} & {mSu} & {Rea} & {SumE} & {OpQA} & {OkVQA} & {Avg} \\
\#Ref & Task: & {\cellcolor[HTML]{FFFFFF}} MT & {\cellcolor[HTML]{FFFFFF}} DTG & {\cellcolor[HTML]{FFFFFF}} Sim & {\cellcolor[HTML]{FFFFFF}} Sim & {\cellcolor[HTML]{FFFFFF}} ImCa & {\cellcolor[HTML]{FFFFFF}} ImCa & {\cellcolor[HTML]{FFFFFF}} Sum & {\cellcolor[HTML]{FFFFFF}} Sum & {\cellcolor[HTML]{FFFFFF}} Sum & {\cellcolor[HTML]{FFFFFF}} QA & {\cellcolor[HTML]{FFFFFF}} VQA & {\cellcolor[HTML]{FFFFFF}} All \\
\midrule
\multirow[c]{10}{*}{Max} & ROUGE-1 & {\cellcolor[HTML]{E25508}} 16.0 & {\cellcolor[HTML]{C14002}} 63.6 & {\cellcolor[HTML]{CE4401}} 61.8 & {\cellcolor[HTML]{E15307}} 41.6 & {\cellcolor[HTML]{9A3103}} 52.3 & {\cellcolor[HTML]{C34002}} 48.7 & {\cellcolor[HTML]{C14002}} 50.7 & {\cellcolor[HTML]{A43503}} 47.4 & {\cellcolor[HTML]{FD9446}} 16.7 & {\cellcolor[HTML]{7F2704}} 35.5 & {\cellcolor[HTML]{862A04}} 19.3 & {\cellcolor[HTML]{802704}} 14.2 \\
 & ROUGE-L & {\cellcolor[HTML]{DB4B03}} 16.8 & {\cellcolor[HTML]{CE4401}} 60.9 & {\cellcolor[HTML]{DA4902}} 59.4 & {\cellcolor[HTML]{E35608}} 40.9 & {\cellcolor[HTML]{9B3203}} 52.0 & {\cellcolor[HTML]{BE3F02}} 49.4 & {\cellcolor[HTML]{B33B02}} 52.9 & {\cellcolor[HTML]{C64102}} 42.6 & {\cellcolor[HTML]{FDA660}} 14.2 & {\cellcolor[HTML]{7F2704}} 35.4 & {\cellcolor[HTML]{7F2704}} 19.8 & {\cellcolor[HTML]{882A04}} 13.9 \\
 & BLEU & {\cellcolor[HTML]{E65A0B}} 15.5 & {\cellcolor[HTML]{CD4401}} 61.3 & {\cellcolor[HTML]{F4711C}} 47.6 & {\cellcolor[HTML]{F87D29}} 32.7 & {\cellcolor[HTML]{A53603}} 50.3 & {\cellcolor[HTML]{AD3803}} 52.0 & {\cellcolor[HTML]{D34601}} 48.0 & {\cellcolor[HTML]{E25508}} 37.6 & {\cellcolor[HTML]{FDB97D}} 11.7 & {\cellcolor[HTML]{FDC28B}} 10.8 & {\cellcolor[HTML]{CB4302}} 15.5 & {\cellcolor[HTML]{B53B02}} 12.0 \\
 & METEOR & {\cellcolor[HTML]{E15307}} 16.2 & {\cellcolor[HTML]{C03F02}} 63.7 & {\cellcolor[HTML]{BB3D02}} 65.8 & {\cellcolor[HTML]{E4580A}} 40.6 & {\cellcolor[HTML]{852904}} 56.0 & {\cellcolor[HTML]{942F03}} 56.5 & {\cellcolor[HTML]{DB4B03}} 46.3 & {\cellcolor[HTML]{7F2704}} 53.7 & {\cellcolor[HTML]{FD9040}} 17.3 & {\cellcolor[HTML]{8E2D04}} 33.7 & {\cellcolor[HTML]{FDCFA0}} 5.1 & {\cellcolor[HTML]{802704}} 14.2 \\
 & BERTScore P & {\cellcolor[HTML]{A53603}} 20.0 & {\cellcolor[HTML]{D04501}} 60.6 & {\cellcolor[HTML]{A63603}} 69.9 & {\cellcolor[HTML]{ED6310}} 37.9 & {\cellcolor[HTML]{A93703}} 49.6 & {\cellcolor[HTML]{C64102}} 48.2 & {\cellcolor[HTML]{842904}} 61.6 & {\cellcolor[HTML]{F9802D}} 29.3 & {\cellcolor[HTML]{FDCE9E}} 9.1 & {\cellcolor[HTML]{FDBD83}} 11.5 & {\cellcolor[HTML]{FDD1A4}} 4.8 & {\cellcolor[HTML]{A43503}} 12.6 \\
 & BERTScore R & {\cellcolor[HTML]{A53603}} 20.0 & {\cellcolor[HTML]{952F03}} 72.9 & {\cellcolor[HTML]{993103}} 73.3 & {\cellcolor[HTML]{F26B15}} 36.3 & {\cellcolor[HTML]{932F03}} 53.5 & {\cellcolor[HTML]{E75B0B}} 41.1 & {\cellcolor[HTML]{8A2B04}} 60.4 & {\cellcolor[HTML]{B13A03}} 45.4 & {\cellcolor[HTML]{FDA35C}} 14.7 & {\cellcolor[HTML]{FDB06E}} 13.1 & {\cellcolor[HTML]{C34002}} 15.9 & {\cellcolor[HTML]{852904}} 14.0 \\
 & BERTScore F1 & {\cellcolor[HTML]{9E3303}} 20.5 & {\cellcolor[HTML]{C84202}} 62.1 & {\cellcolor[HTML]{973003}} 73.4 & {\cellcolor[HTML]{EE6511}} 37.5 & {\cellcolor[HTML]{973003}} 52.6 & {\cellcolor[HTML]{CB4302}} 47.4 & {\cellcolor[HTML]{7F2704}} 62.8 & {\cellcolor[HTML]{DC4C03}} 39.3 & {\cellcolor[HTML]{FDAF6C}} 13.1 & {\cellcolor[HTML]{FDB576}} 12.4 & {\cellcolor[HTML]{FD8C3B}} 9.9 & {\cellcolor[HTML]{902E04}} 13.5 \\
 & BLEURT & {\cellcolor[HTML]{7F2704}} 22.8 & {\cellcolor[HTML]{A83703}} 68.4 & {\cellcolor[HTML]{7F2704}} 79.9 & {\cellcolor[HTML]{EE6410}} 37.7 & {\cellcolor[HTML]{7F2704}} 57.3 & {\cellcolor[HTML]{7F2704}} 60.6 & {\cellcolor[HTML]{DB4B03}} 46.3 & {\cellcolor[HTML]{EF6612}} 34.1 & {\cellcolor[HTML]{FDCB9B}} 9.4 & {\cellcolor[HTML]{EF6612}} 22.6 & {\cellcolor[HTML]{952F03}} 18.4 & {\cellcolor[HTML]{7F2704}} 14.3 \\
 & Nubia & {\cellcolor[HTML]{882A04}} 22.1 & {\cellcolor[HTML]{7F2704}} 78.7 & {\cellcolor[HTML]{CD4401}} 62.2 & {\cellcolor[HTML]{DB4A02}} 43.5 & {\cellcolor[HTML]{963003}} 52.9 & {\cellcolor[HTML]{892B04}} 58.6 & {\cellcolor[HTML]{F3701B}} 37.7 & {\cellcolor[HTML]{FDD3A9}} 12.5 & {\cellcolor[HTML]{FEDEBF}} 6.0 & {\cellcolor[HTML]{902E04}} 33.5 & {\cellcolor[HTML]{E35608}} 13.8 & {\cellcolor[HTML]{963003}} 13.2 \\
\midrule
\multirow[c]{10}{*}{1} & ROUGE-1 & {\cellcolor[HTML]{E25508}} 16.0 & {\cellcolor[HTML]{A23503}} 69.7 & {\cellcolor[HTML]{F3701B}} 47.9 & {\cellcolor[HTML]{E15307}} 41.6 & {\cellcolor[HTML]{D64701}} 43.4 & {\cellcolor[HTML]{F26D17}} 37.1 & {\cellcolor[HTML]{C14002}} 50.7 & {\cellcolor[HTML]{A43503}} 47.4 & {\cellcolor[HTML]{FC8A39}} 17.9 & {\cellcolor[HTML]{7F2704}} 35.5 & {\cellcolor[HTML]{832804}} 19.5 & {\cellcolor[HTML]{942F03}} 13.3 \\
 & ROUGE-L & {\cellcolor[HTML]{DB4B03}} 16.8 & {\cellcolor[HTML]{CD4401}} 61.2 & {\cellcolor[HTML]{F98230}} 43.0 & {\cellcolor[HTML]{E35608}} 40.9 & {\cellcolor[HTML]{DE5005}} 41.4 & {\cellcolor[HTML]{F06712}} 38.2 & {\cellcolor[HTML]{B33B02}} 52.9 & {\cellcolor[HTML]{C64102}} 42.6 & {\cellcolor[HTML]{FD9B50}} 15.7 & {\cellcolor[HTML]{7F2704}} 35.4 & {\cellcolor[HTML]{832804}} 19.5 & {\cellcolor[HTML]{A13403}} 12.7 \\
 & BLEU & {\cellcolor[HTML]{E65A0B}} 15.5 & {\cellcolor[HTML]{E65A0B}} 53.6 & {\cellcolor[HTML]{FDAF6C}} 29.9 & {\cellcolor[HTML]{F87D29}} 32.7 & {\cellcolor[HTML]{FC8937}} 29.5 & {\cellcolor[HTML]{FA8331}} 32.2 & {\cellcolor[HTML]{D34601}} 48.0 & {\cellcolor[HTML]{E25508}} 37.6 & {\cellcolor[HTML]{FDD9B5}} 7.0 & {\cellcolor[HTML]{FDC28B}} 10.8 & {\cellcolor[HTML]{8B2C04}} 19.0 & {\cellcolor[HTML]{E45709}} 9.9 \\
 & METEOR & {\cellcolor[HTML]{E15307}} 16.2 & {\cellcolor[HTML]{AB3803}} 67.9 & {\cellcolor[HTML]{EB610F}} 52.2 & {\cellcolor[HTML]{E4580A}} 40.6 & {\cellcolor[HTML]{D94801}} 42.9 & {\cellcolor[HTML]{E5590A}} 41.6 & {\cellcolor[HTML]{DB4B03}} 46.3 & {\cellcolor[HTML]{7F2704}} 53.7 & {\cellcolor[HTML]{FD984B}} 16.2 & {\cellcolor[HTML]{8E2D04}} 33.7 & {\cellcolor[HTML]{FDC692}} 5.7 & {\cellcolor[HTML]{9B3203}} 13.0 \\
 & BERTScore P & {\cellcolor[HTML]{A53603}} 20.0 & {\cellcolor[HTML]{D84801}} 59.2 & {\cellcolor[HTML]{F67824}} 45.8 & {\cellcolor[HTML]{ED6310}} 37.9 & {\cellcolor[HTML]{EC620F}} 37.3 & {\cellcolor[HTML]{F36E19}} 36.7 & {\cellcolor[HTML]{842904}} 61.6 & {\cellcolor[HTML]{F9802D}} 29.3 & {\cellcolor[HTML]{FDCE9E}} 9.1 & {\cellcolor[HTML]{FDBD83}} 11.5 & {\cellcolor[HTML]{FEDCBB}} 3.6 & {\cellcolor[HTML]{D14501}} 11.0 \\
 & BERTScore R & {\cellcolor[HTML]{A53603}} 20.0 & {\cellcolor[HTML]{9E3303}} 70.8 & {\cellcolor[HTML]{B83C02}} 66.3 & {\cellcolor[HTML]{F26B15}} 36.3 & {\cellcolor[HTML]{C34002}} 45.9 & {\cellcolor[HTML]{FDA35C}} 25.2 & {\cellcolor[HTML]{8A2B04}} 60.4 & {\cellcolor[HTML]{B13A03}} 45.4 & {\cellcolor[HTML]{FDA762}} 14.1 & {\cellcolor[HTML]{FDB06E}} 13.1 & {\cellcolor[HTML]{FDAF6C}} 7.4 & {\cellcolor[HTML]{A13403}} 12.7 \\
 & BERTScore F1 & {\cellcolor[HTML]{9E3303}} 20.5 & {\cellcolor[HTML]{D04501}} 60.8 & {\cellcolor[HTML]{D14501}} 61.4 & {\cellcolor[HTML]{EE6511}} 37.5 & {\cellcolor[HTML]{D14501}} 43.9 & {\cellcolor[HTML]{F87E2B}} 33.5 & {\cellcolor[HTML]{7F2704}} 62.8 & {\cellcolor[HTML]{DC4C03}} 39.3 & {\cellcolor[HTML]{FDB475}} 12.4 & {\cellcolor[HTML]{FDB576}} 12.4 & {\cellcolor[HTML]{FDBF86}} 6.2 & {\cellcolor[HTML]{AE3903}} 12.2 \\
 & BLEURT & {\cellcolor[HTML]{7F2704}} 22.8 & {\cellcolor[HTML]{852904}} 77.1 & {\cellcolor[HTML]{AE3903}} 68.1 & {\cellcolor[HTML]{EE6410}} 37.7 & {\cellcolor[HTML]{9E3303}} 51.6 & {\cellcolor[HTML]{A53603}} 53.2 & {\cellcolor[HTML]{DB4B03}} 46.3 & {\cellcolor[HTML]{EF6612}} 34.1 & {\cellcolor[HTML]{FDC895}} 9.8 & {\cellcolor[HTML]{EF6612}} 22.6 & {\cellcolor[HTML]{D14501}} 15.2 & {\cellcolor[HTML]{8B2C04}} 13.7 \\
 & Nubia & {\cellcolor[HTML]{882A04}} 22.1 & {\cellcolor[HTML]{7F2704}} 78.7 & {\cellcolor[HTML]{CD4401}} 62.2 & {\cellcolor[HTML]{DB4A02}} 43.5 & {\cellcolor[HTML]{963003}} 52.9 & {\cellcolor[HTML]{892B04}} 58.6 & {\cellcolor[HTML]{F3701B}} 37.7 & {\cellcolor[HTML]{FDD3A9}} 12.5 & {\cellcolor[HTML]{FEDEBF}} 6.0 & {\cellcolor[HTML]{902E04}} 33.5 & {\cellcolor[HTML]{E35608}} 13.8 & {\cellcolor[HTML]{963003}} 13.2 \\
\midrule
\multirow[c]{8}{*}{0} & Abstr-1 & {\cellcolor[HTML]{FFEEDD}} 1.4 & {\cellcolor[HTML]{FFF5EB}} - & {\cellcolor[HTML]{FEDCBB}} -14.4 & {\cellcolor[HTML]{7F2704}} -58.8 & {\cellcolor[HTML]{FFF5EB}} - & {\cellcolor[HTML]{FFF5EB}} - & {\cellcolor[HTML]{FFF5EB}} - & {\cellcolor[HTML]{FEDFC0}} 8.9 & {\cellcolor[HTML]{F87F2C}} -19.3 & {\cellcolor[HTML]{FDD1A4}} -8.7 & {\cellcolor[HTML]{FDC38D}} 5.9 & {\cellcolor[HTML]{FDDBB8}} -2.7 \\
 & Abstr-3 & {\cellcolor[HTML]{FFEEDE}} 1.3 & {\cellcolor[HTML]{FFF5EB}} - & {\cellcolor[HTML]{FDA660}} -32.2 & {\cellcolor[HTML]{8F2D04}} -55.7 & {\cellcolor[HTML]{FFF5EB}} - & {\cellcolor[HTML]{FFF5EB}} - & {\cellcolor[HTML]{FFF5EB}} - & {\cellcolor[HTML]{FFEFDF}} -2.8 & {\cellcolor[HTML]{7F2704}} -35.2 & {\cellcolor[HTML]{FDAE6A}} 13.4 & {\cellcolor[HTML]{FDA762}} 7.9 & {\cellcolor[HTML]{FDD5AB}} -3.2 \\
 & Length & {\cellcolor[HTML]{FEECDA}} 1.7 & {\cellcolor[HTML]{FDD1A3}} 19.4 & {\cellcolor[HTML]{FDD1A4}} 19.4 & {\cellcolor[HTML]{FDC794}} 16.6 & {\cellcolor[HTML]{FDCD9C}} 15.1 & {\cellcolor[HTML]{FDCB9B}} -16.2 & {\cellcolor[HTML]{FEEDDC}} -4.4 & {\cellcolor[HTML]{F87F2C}} 29.4 & {\cellcolor[HTML]{FDD4AA}} 8.1 & {\cellcolor[HTML]{FDA25A}} 14.9 & {\cellcolor[HTML]{FD9446}} -9.4 & {\cellcolor[HTML]{FDD7B1}} 3.0 \\
 & Repet-1 & {\cellcolor[HTML]{FEECDA}} 1.7 & {\cellcolor[HTML]{FDD3A7}} 18.5 & {\cellcolor[HTML]{FDD1A4}} 19.4 & {\cellcolor[HTML]{FDC794}} 16.6 & {\cellcolor[HTML]{FDCFA0}} 14.6 & {\cellcolor[HTML]{FDCB9B}} -16.2 & {\cellcolor[HTML]{FEEDDC}} -4.4 & {\cellcolor[HTML]{F87F2C}} 29.4 & {\cellcolor[HTML]{FDD4AA}} 8.1 & {\cellcolor[HTML]{FDA25A}} 14.9 & {\cellcolor[HTML]{FD9446}} -9.4 & {\cellcolor[HTML]{FDD9B4}} 2.9 \\
 & Repet-3 & {\cellcolor[HTML]{FEECDA}} 1.7 & {\cellcolor[HTML]{FDD7AF}} 16.8 & {\cellcolor[HTML]{FDD1A4}} 19.4 & {\cellcolor[HTML]{FDC794}} 16.6 & {\cellcolor[HTML]{FDCD9C}} 15.0 & {\cellcolor[HTML]{FDCB9B}} -16.2 & {\cellcolor[HTML]{FEEDDC}} -4.3 & {\cellcolor[HTML]{F87F2C}} 29.4 & {\cellcolor[HTML]{FDD4AA}} 8.1 & {\cellcolor[HTML]{FDB271}} 12.8 & {\cellcolor[HTML]{FD8F3E}} -9.8 & {\cellcolor[HTML]{FDD9B5}} 2.8 \\
 & -GPT2 Perpl. & {\cellcolor[HTML]{FEE1C4}} 3.5 & {\cellcolor[HTML]{FDBF86}} 24.7 & {\cellcolor[HTML]{FDC590}} 23.1 & {\cellcolor[HTML]{FEDDBC}} 10.5 & {\cellcolor[HTML]{FDD0A2}} 14.4 & {\cellcolor[HTML]{FFEFE0}} -2.9 & {\cellcolor[HTML]{FFEFE0}} 3.0 & {\cellcolor[HTML]{FDD3A7}} 12.6 & {\cellcolor[HTML]{FEEBD7}} -3.1 & {\cellcolor[HTML]{FFF2E5}} 1.1 & {\cellcolor[HTML]{FD9E54}} 8.6 & {\cellcolor[HTML]{FDD7B1}} 3.0 \\
\bottomrule
\end{tabular}

\caption{\textbf{Correctness dimension}: Pearson coefficient between automatic metrics and human judgement for Correctness on the 10 human evaluation datasets. The top bloc corresponds to coefficients computed when all the human references were available. The second bloc corresponds to coefficients computed given a single human reference. The third bloc corresponds to coefficients computed given no human reference.}
\vspace{-0.5cm}
\label{tab:main_res_correctness}
\end{table}

%% file: main_table_other_pearson.tex
\begin{table}[h]
\tiny

\begin{tabular}{llccccccccccc}
\toprule
{} & {Dataset:} & {Web} & {Asv} & {Asv} & {MUS} & {mSu} & {SumE} & {SumE} & {SumE} & {OpQA} & {OkVQA} & {OkVQA} \\
\multirow[c]{2}{*}{\#Ref} & Task: & {\cellcolor[HTML]{FFFFFF}} DTG & {\cellcolor[HTML]{FFFFFF}} Sim & {\cellcolor[HTML]{FFFFFF}} Sim & {\cellcolor[HTML]{FFFFFF}} Sim & {\cellcolor[HTML]{FFFFFF}} Sum & {\cellcolor[HTML]{FFFFFF}} Sum & {\cellcolor[HTML]{FFFFFF}} Sum & {\cellcolor[HTML]{FFFFFF}} Sum & {\cellcolor[HTML]{FFFFFF}} QA & {\cellcolor[HTML]{FFFFFF}} VQA & {\cellcolor[HTML]{FFFFFF}} VQA \\
 & Dim: & {\cellcolor[HTML]{FFFFFF}} Flu & {\cellcolor[HTML]{FFFFFF}} Flu & {\cellcolor[HTML]{FFFFFF}} Sim & {\cellcolor[HTML]{FFFFFF}} Flu & {\cellcolor[HTML]{FFFFFF}} Rel & {\cellcolor[HTML]{FFFFFF}} Rel & {\cellcolor[HTML]{FFFFFF}} Coh & {\cellcolor[HTML]{FFFFFF}} Flu & {\cellcolor[HTML]{FFFFFF}} Obv & {\cellcolor[HTML]{FFFFFF}} Pos & {\cellcolor[HTML]{FFFFFF}} Obv \\
\midrule
\multirow[c]{10}{*}{Max} & ROUGE-1 & {\cellcolor[HTML]{C94202}} 51.0 & {\cellcolor[HTML]{EC620F}} 42.0 & {\cellcolor[HTML]{DC4C03}} 42.4 & {\cellcolor[HTML]{BB3D02}} 26.1 & {\cellcolor[HTML]{CB4302}} 50.9 & {\cellcolor[HTML]{C03F02}} 32.3 & {\cellcolor[HTML]{FB8735}} 18.1 & {\cellcolor[HTML]{FC8B3A}} 13.6 & {\cellcolor[HTML]{7F2704}} 41.7 & {\cellcolor[HTML]{FD974A}} 14.3 & {\cellcolor[HTML]{832804}} 29.1 \\
 & ROUGE-L & {\cellcolor[HTML]{BD3E02}} 52.9 & {\cellcolor[HTML]{EF6612}} 40.9 & {\cellcolor[HTML]{E15307}} 41.0 & {\cellcolor[HTML]{C54102}} 25.4 & {\cellcolor[HTML]{C03F02}} 52.8 & {\cellcolor[HTML]{F16913}} 24.8 & {\cellcolor[HTML]{FD9B50}} 15.5 & {\cellcolor[HTML]{FDA45D}} 11.0 & {\cellcolor[HTML]{7F2704}} 41.6 & {\cellcolor[HTML]{FD974A}} 14.3 & {\cellcolor[HTML]{7F2704}} 29.5 \\
 & BLEU & {\cellcolor[HTML]{B53B02}} 54.2 & {\cellcolor[HTML]{FD9B50}} 28.9 & {\cellcolor[HTML]{FC8A39}} 29.5 & {\cellcolor[HTML]{E15407}} 22.4 & {\cellcolor[HTML]{D34601}} 49.6 & {\cellcolor[HTML]{DD4D04}} 29.0 & {\cellcolor[HTML]{DE5005}} 25.0 & {\cellcolor[HTML]{FB8836}} 13.9 & {\cellcolor[HTML]{FDC692}} 12.0 & {\cellcolor[HTML]{FDD3A9}} 7.2 & {\cellcolor[HTML]{E35608}} 20.6 \\
 & METEOR & {\cellcolor[HTML]{CD4401}} 50.3 & {\cellcolor[HTML]{EE6511}} 41.3 & {\cellcolor[HTML]{E95E0D}} 38.5 & {\cellcolor[HTML]{B53B02}} 26.6 & {\cellcolor[HTML]{B13A03}} 54.9 & {\cellcolor[HTML]{9E3303}} 36.0 & {\cellcolor[HTML]{FD9243}} 16.8 & {\cellcolor[HTML]{FDA159}} 11.4 & {\cellcolor[HTML]{8A2B04}} 40.1 & {\cellcolor[HTML]{FFF5EA}} 0.2 & {\cellcolor[HTML]{FEE6CF}} 3.6 \\
 & BERTScore P & {\cellcolor[HTML]{932F03}} 60.6 & {\cellcolor[HTML]{A43503}} 57.0 & {\cellcolor[HTML]{832804}} 57.1 & {\cellcolor[HTML]{EA5F0E}} 21.1 & {\cellcolor[HTML]{C03F02}} 52.6 & {\cellcolor[HTML]{F26B15}} 24.7 & {\cellcolor[HTML]{A03403}} 31.0 & {\cellcolor[HTML]{F36E19}} 16.3 & {\cellcolor[HTML]{FDD0A2}} 10.5 & {\cellcolor[HTML]{FD994D}} 14.1 & {\cellcolor[HTML]{F26C16}} 18.2 \\
 & BERTScore R & {\cellcolor[HTML]{7F2704}} 64.7 & {\cellcolor[HTML]{A43503}} 56.9 & {\cellcolor[HTML]{A93703}} 50.1 & {\cellcolor[HTML]{E95E0D}} 21.2 & {\cellcolor[HTML]{7F2704}} 65.0 & {\cellcolor[HTML]{7F2704}} 39.9 & {\cellcolor[HTML]{862A04}} 33.8 & {\cellcolor[HTML]{FC8937}} 13.8 & {\cellcolor[HTML]{FDCB9B}} 11.1 & {\cellcolor[HTML]{FDA057}} 13.2 & {\cellcolor[HTML]{BB3D02}} 24.3 \\
 & BERTScore F1 & {\cellcolor[HTML]{942F03}} 60.3 & {\cellcolor[HTML]{9F3303}} 58.0 & {\cellcolor[HTML]{8F2D04}} 54.7 & {\cellcolor[HTML]{E75C0C}} 21.4 & {\cellcolor[HTML]{952F03}} 60.3 & {\cellcolor[HTML]{AE3903}} 34.0 & {\cellcolor[HTML]{7F2704}} 34.7 & {\cellcolor[HTML]{F16913}} 16.7 & {\cellcolor[HTML]{FDCE9E}} 10.9 & {\cellcolor[HTML]{FD9547}} 14.6 & {\cellcolor[HTML]{D34601}} 22.5 \\
 & BLEURT & {\cellcolor[HTML]{A93703}} 56.1 & {\cellcolor[HTML]{7F2704}} 64.6 & {\cellcolor[HTML]{7F2704}} 57.8 & {\cellcolor[HTML]{7F2704}} 31.8 & {\cellcolor[HTML]{E85D0C}} 43.5 & {\cellcolor[HTML]{E5590A}} 27.3 & {\cellcolor[HTML]{FDA55F}} 14.2 & {\cellcolor[HTML]{EE6410}} 17.2 & {\cellcolor[HTML]{FC8A39}} 21.2 & {\cellcolor[HTML]{7F2704}} 30.9 & {\cellcolor[HTML]{932F03}} 27.6 \\
 & Nubia & {\cellcolor[HTML]{CD4401}} 50.4 & {\cellcolor[HTML]{E75C0C}} 43.6 & {\cellcolor[HTML]{E65A0B}} 39.3 & {\cellcolor[HTML]{D34601}} 24.3 & {\cellcolor[HTML]{FD9A4E}} 29.4 & {\cellcolor[HTML]{FDB271}} 14.4 & {\cellcolor[HTML]{FDD7B1}} 7.2 & {\cellcolor[HTML]{FDC895}} 7.5 & {\cellcolor[HTML]{993103}} 38.2 & {\cellcolor[HTML]{FDC692}} 8.9 & {\cellcolor[HTML]{E5590A}} 20.2 \\
\midrule
\multirow[c]{10}{*}{1} & ROUGE-1 & {\cellcolor[HTML]{AD3803}} 55.4 & {\cellcolor[HTML]{FB8735}} 33.7 & {\cellcolor[HTML]{F9812E}} 31.2 & {\cellcolor[HTML]{BB3D02}} 26.1 & {\cellcolor[HTML]{CB4302}} 50.9 & {\cellcolor[HTML]{B63C02}} 33.3 & {\cellcolor[HTML]{F9812E}} 18.8 & {\cellcolor[HTML]{FC8B3A}} 13.6 & {\cellcolor[HTML]{7F2704}} 41.7 & {\cellcolor[HTML]{FD9B50}} 13.8 & {\cellcolor[HTML]{852904}} 28.9 \\
 & ROUGE-L & {\cellcolor[HTML]{C34002}} 52.0 & {\cellcolor[HTML]{FD8F3E}} 31.8 & {\cellcolor[HTML]{FD8F3E}} 28.5 & {\cellcolor[HTML]{C54102}} 25.4 & {\cellcolor[HTML]{C03F02}} 52.8 & {\cellcolor[HTML]{E85D0C}} 26.8 & {\cellcolor[HTML]{FB8634}} 18.2 & {\cellcolor[HTML]{FD9A4E}} 12.1 & {\cellcolor[HTML]{7F2704}} 41.6 & {\cellcolor[HTML]{FD9B50}} 13.8 & {\cellcolor[HTML]{852904}} 28.9 \\
 & BLEU & {\cellcolor[HTML]{E75B0B}} 43.8 & {\cellcolor[HTML]{FDA863}} 25.6 & {\cellcolor[HTML]{FDA55F}} 23.5 & {\cellcolor[HTML]{E15407}} 22.4 & {\cellcolor[HTML]{D34601}} 49.6 & {\cellcolor[HTML]{FB8735}} 20.8 & {\cellcolor[HTML]{FDAF6C}} 13.0 & {\cellcolor[HTML]{FDCFA0}} 6.9 & {\cellcolor[HTML]{FDC692}} 12.0 & {\cellcolor[HTML]{FDBF86}} 9.7 & {\cellcolor[HTML]{C64102}} 23.5 \\
 & METEOR & {\cellcolor[HTML]{B83C02}} 53.8 & {\cellcolor[HTML]{F9802D}} 35.3 & {\cellcolor[HTML]{F87F2C}} 31.7 & {\cellcolor[HTML]{B53B02}} 26.6 & {\cellcolor[HTML]{B13A03}} 54.9 & {\cellcolor[HTML]{D34601}} 30.4 & {\cellcolor[HTML]{FD9D53}} 15.3 & {\cellcolor[HTML]{FD9D53}} 11.8 & {\cellcolor[HTML]{8A2B04}} 40.1 & {\cellcolor[HTML]{FFF1E4}} 1.0 & {\cellcolor[HTML]{FEE2C7}} 4.3 \\
 & BERTScore P & {\cellcolor[HTML]{9C3203}} 58.6 & {\cellcolor[HTML]{F36E19}} 39.2 & {\cellcolor[HTML]{E05206}} 41.2 & {\cellcolor[HTML]{EA5F0E}} 21.1 & {\cellcolor[HTML]{C03F02}} 52.6 & {\cellcolor[HTML]{E95E0D}} 26.5 & {\cellcolor[HTML]{C64102}} 27.6 & {\cellcolor[HTML]{FB8836}} 13.9 & {\cellcolor[HTML]{FDD0A2}} 10.5 & {\cellcolor[HTML]{FD9A4E}} 13.9 & {\cellcolor[HTML]{F98230}} 15.9 \\
 & BERTScore R & {\cellcolor[HTML]{8F2D04}} 61.4 & {\cellcolor[HTML]{D14501}} 49.5 & {\cellcolor[HTML]{CD4401}} 45.1 & {\cellcolor[HTML]{E95E0D}} 21.2 & {\cellcolor[HTML]{7F2704}} 65.0 & {\cellcolor[HTML]{993103}} 36.5 & {\cellcolor[HTML]{B13A03}} 29.4 & {\cellcolor[HTML]{FD9344}} 12.8 & {\cellcolor[HTML]{FDCB9B}} 11.1 & {\cellcolor[HTML]{FDC28B}} 9.3 & {\cellcolor[HTML]{FD8F3E}} 14.6 \\
 & BERTScore F1 & {\cellcolor[HTML]{9C3203}} 58.5 & {\cellcolor[HTML]{D84801}} 48.5 & {\cellcolor[HTML]{C03F02}} 46.8 & {\cellcolor[HTML]{E75C0C}} 21.4 & {\cellcolor[HTML]{952F03}} 60.3 & {\cellcolor[HTML]{AE3903}} 34.0 & {\cellcolor[HTML]{A23503}} 30.7 & {\cellcolor[HTML]{F9812E}} 14.5 & {\cellcolor[HTML]{FDCE9E}} 10.9 & {\cellcolor[HTML]{FD9E54}} 13.5 & {\cellcolor[HTML]{F5741F}} 17.3 \\
 & BLEURT & {\cellcolor[HTML]{812804}} 64.0 & {\cellcolor[HTML]{AD3803}} 55.3 & {\cellcolor[HTML]{B33B02}} 48.7 & {\cellcolor[HTML]{7F2704}} 31.8 & {\cellcolor[HTML]{E85D0C}} 43.5 & {\cellcolor[HTML]{E15407}} 28.1 & {\cellcolor[HTML]{FDA35C}} 14.4 & {\cellcolor[HTML]{F87E2B}} 14.8 & {\cellcolor[HTML]{FC8A39}} 21.2 & {\cellcolor[HTML]{A93703}} 26.7 & {\cellcolor[HTML]{C54102}} 23.6 \\
 & Nubia & {\cellcolor[HTML]{CD4401}} 50.4 & {\cellcolor[HTML]{E75C0C}} 43.6 & {\cellcolor[HTML]{E65A0B}} 39.3 & {\cellcolor[HTML]{D34601}} 24.3 & {\cellcolor[HTML]{FD9A4E}} 29.4 & {\cellcolor[HTML]{FDB271}} 14.4 & {\cellcolor[HTML]{FDD7B1}} 7.2 & {\cellcolor[HTML]{FDC895}} 7.5 & {\cellcolor[HTML]{993103}} 38.2 & {\cellcolor[HTML]{FDC692}} 8.9 & {\cellcolor[HTML]{E5590A}} 20.2 \\
\midrule
\multirow[c]{8}{*}{0} & Abstr-1 & {\cellcolor[HTML]{FFF5EB}} - & {\cellcolor[HTML]{FEE0C3}} -10.2 & {\cellcolor[HTML]{FEE5CC}} -7.5 & {\cellcolor[HTML]{852904}} -31.1 & {\cellcolor[HTML]{FFF5EB}} - & {\cellcolor[HTML]{F26C16}} -24.5 & {\cellcolor[HTML]{CB4302}} -27.2 & {\cellcolor[HTML]{FD8E3D}} -13.3 & {\cellcolor[HTML]{FEE5CC}} -5.4 & {\cellcolor[HTML]{FDB170}} 11.3 & {\cellcolor[HTML]{FDD0A2}} 7.4 \\
 & Abstr-3 & {\cellcolor[HTML]{FFF5EB}} - & {\cellcolor[HTML]{FDB576}} -22.7 & {\cellcolor[HTML]{FDD1A3}} -14.3 & {\cellcolor[HTML]{B83C02}} -26.4 & {\cellcolor[HTML]{FFF5EB}} - & {\cellcolor[HTML]{CE4401}} -31.0 & {\cellcolor[HTML]{A43503}} -30.5 & {\cellcolor[HTML]{7F2704}} -26.8 & {\cellcolor[HTML]{FDD4AA}} 9.6 & {\cellcolor[HTML]{FDC590}} 9.0 & {\cellcolor[HTML]{FEECDA}} -2.3 \\
 & Length & {\cellcolor[HTML]{FFF2E6}} -1.7 & {\cellcolor[HTML]{FFF1E3}} 2.5 & {\cellcolor[HTML]{FFF4E8}} -0.8 & {\cellcolor[HTML]{FDD6AE}} 6.9 & {\cellcolor[HTML]{FEE0C1}} 10.5 & {\cellcolor[HTML]{E95E0D}} 26.6 & {\cellcolor[HTML]{FDD1A3}} 8.6 & {\cellcolor[HTML]{FEE8D2}} -2.9 & {\cellcolor[HTML]{FDCD9C}} 11.0 & {\cellcolor[HTML]{F26C16}} -19.0 & {\cellcolor[HTML]{F87D29}} -16.4 \\
 & Repet-1 & {\cellcolor[HTML]{FFF0E1}} -2.8 & {\cellcolor[HTML]{FFF1E3}} 2.5 & {\cellcolor[HTML]{FFF4E8}} -0.8 & {\cellcolor[HTML]{FDD6AE}} 6.9 & {\cellcolor[HTML]{FEE0C1}} 10.5 & {\cellcolor[HTML]{E95E0D}} 26.6 & {\cellcolor[HTML]{FDD1A3}} 8.6 & {\cellcolor[HTML]{FEE8D2}} -2.9 & {\cellcolor[HTML]{FDCD9C}} 11.0 & {\cellcolor[HTML]{F26C16}} -19.0 & {\cellcolor[HTML]{F87D29}} -16.4 \\
 & Repet-3 & {\cellcolor[HTML]{FEECDA}} -5.0 & {\cellcolor[HTML]{FFF1E3}} 2.5 & {\cellcolor[HTML]{FFF4E8}} -0.8 & {\cellcolor[HTML]{FDD6AE}} 6.9 & {\cellcolor[HTML]{FEE0C1}} 10.6 & {\cellcolor[HTML]{E95E0D}} 26.6 & {\cellcolor[HTML]{FDD1A3}} 8.6 & {\cellcolor[HTML]{FEE8D2}} -2.9 & {\cellcolor[HTML]{FDD1A4}} 10.2 & {\cellcolor[HTML]{EE6511}} -19.7 & {\cellcolor[HTML]{F77A27}} -16.6 \\
 & -GPT2 Perpl. & {\cellcolor[HTML]{FDD0A2}} 16.2 & {\cellcolor[HTML]{FDB475}} 22.9 & {\cellcolor[HTML]{FDB475}} 20.4 & {\cellcolor[HTML]{FDAD69}} 12.1 & {\cellcolor[HTML]{FFF2E5}} 1.9 & {\cellcolor[HTML]{FD8E3D}} 19.8 & {\cellcolor[HTML]{FD9A4E}} 15.7 & {\cellcolor[HTML]{FDB97D}} 8.9 & {\cellcolor[HTML]{FFF5EA}} 0.2 & {\cellcolor[HTML]{FEE0C3}} -4.9 & {\cellcolor[HTML]{FEE1C4}} 4.6 \\
\bottomrule
\end{tabular}

\caption{\textbf{Non Correctness dimensions}: Pearson coefficient between automatic metrics and human judgement for the dimensions other than Correctness. The top bloc corresponds to coefficients computed when all the human references were available. The second bloc corresponds to coefficients computed given a single human reference. The third bloc corresponds to coefficients computed given no human reference.}
\vspace{-0.5cm}
\label{tab:main_res_other}
\end{table}

%% file: supp_material/main_table_correctness_kendall.tex
\begin{table}[h]
\tiny

\begin{tabular}{llcccccccccccc}
\toprule
{} & {Dataset:} & {WMT} & {Web} & {Asv} & {MUS} & {Pas} & {Fli} & {mSu} & {Rea} & {SumE} & {OpQA} & {OkVQA} & {Avg} \\
\#Ref & Task: & {\cellcolor[HTML]{FFFFFF}} MT & {\cellcolor[HTML]{FFFFFF}} DTG & {\cellcolor[HTML]{FFFFFF}} Sim & {\cellcolor[HTML]{FFFFFF}} Sim & {\cellcolor[HTML]{FFFFFF}} ImCa & {\cellcolor[HTML]{FFFFFF}} ImCa & {\cellcolor[HTML]{FFFFFF}} Sum & {\cellcolor[HTML]{FFFFFF}} Sum & {\cellcolor[HTML]{FFFFFF}} Sum & {\cellcolor[HTML]{FFFFFF}} QA & {\cellcolor[HTML]{FFFFFF}} VQA & {\cellcolor[HTML]{FFFFFF}} All \\
\midrule
\multirow[c]{10}{*}{Max} & ROUGE-1 & {\cellcolor[HTML]{E65A0B}} 12.6 & {\cellcolor[HTML]{DC4C03}} 44.3 & {\cellcolor[HTML]{E15407}} 43.6 & {\cellcolor[HTML]{C64102}} 33.3 & {\cellcolor[HTML]{9A3103}} 52.2 & {\cellcolor[HTML]{DE5005}} 32.0 & {\cellcolor[HTML]{D54601}} 31.6 & {\cellcolor[HTML]{AD3803}} 32.0 & {\cellcolor[HTML]{FDA159}} 10.5 & {\cellcolor[HTML]{802704}} 32.7 & {\cellcolor[HTML]{D94801}} 11.2 & {\cellcolor[HTML]{8A2B04}} 10.5 \\
 & ROUGE-2 & {\cellcolor[HTML]{EC620F}} 12.0 & {\cellcolor[HTML]{D04501}} 46.7 & {\cellcolor[HTML]{F3701B}} 37.0 & {\cellcolor[HTML]{D54601}} 31.8 & {\cellcolor[HTML]{9B3203}} 52.0 & {\cellcolor[HTML]{DD4D04}} 32.4 & {\cellcolor[HTML]{D64701}} 31.4 & {\cellcolor[HTML]{A53603}} 32.8 & {\cellcolor[HTML]{FDA55F}} 10.1 & {\cellcolor[HTML]{EB600E}} 21.7 & {\cellcolor[HTML]{FDA45D}} 6.2 & {\cellcolor[HTML]{9E3303}} 9.8 \\
 & ROUGE-L & {\cellcolor[HTML]{E15307}} 13.1 & {\cellcolor[HTML]{DE4E05}} 44.0 & {\cellcolor[HTML]{E95E0D}} 41.2 & {\cellcolor[HTML]{BE3F02}} 34.0 & {\cellcolor[HTML]{9B3203}} 52.0 & {\cellcolor[HTML]{E05206}} 31.6 & {\cellcolor[HTML]{D14501}} 31.8 & {\cellcolor[HTML]{C64102}} 29.8 & {\cellcolor[HTML]{FDB06E}} 9.1 & {\cellcolor[HTML]{802704}} 32.7 & {\cellcolor[HTML]{D54601}} 11.4 & {\cellcolor[HTML]{902E04}} 10.3 \\
\bottomrule
\end{tabular}

\caption{\textbf{Correctness dimensions}: Kendall tau between automatic metrics and human judgement for Correctness. The top bloc corresponds to coefficients computed when all the human references were available. The second bloc corresponds to coefficients computed given a single human reference. The third bloc corresponds to coefficients computed given no human reference.}
\label{tab:kendall_values_corectness}
\end{table}

%% file: supp_material/main_table_other_kendall.tex
\begin{table}[h]
\tiny

\begin{tabular}{llccccccccccc}
\toprule
{} & {Dataset:} & {Web} & {Asv} & {Asv} & {MUS} & {mSu} & {SumE} & {SumE} & {SumE} & {OpQA} & {OkVQA} & {OkVQA} \\
\#Ref & Task: & {\cellcolor[HTML]{FFFFFF}} DTG & {\cellcolor[HTML]{FFFFFF}} Sim & {\cellcolor[HTML]{FFFFFF}} Sim & {\cellcolor[HTML]{FFFFFF}} Sim & {\cellcolor[HTML]{FFFFFF}} Sum & {\cellcolor[HTML]{FFFFFF}} Sum & {\cellcolor[HTML]{FFFFFF}} Sum & {\cellcolor[HTML]{FFFFFF}} Sum & {\cellcolor[HTML]{FFFFFF}} QA & {\cellcolor[HTML]{FFFFFF}} VQA & {\cellcolor[HTML]{FFFFFF}} VQA \\
\midrule
\multirow[c]{10}{*}{Max} & ROUGE-1 & {\cellcolor[HTML]{E85D0C}} 33.5 & {\cellcolor[HTML]{F16913}} 30.5 & {\cellcolor[HTML]{E05206}} 32.2 & {\cellcolor[HTML]{A43503}} 17.9 & {\cellcolor[HTML]{DB4B03}} 32.5 & {\cellcolor[HTML]{DE5005}} 20.9 & {\cellcolor[HTML]{FD8E3D}} 12.3 & {\cellcolor[HTML]{FDB576}} 6.2 & {\cellcolor[HTML]{862A04}} 37.2 & {\cellcolor[HTML]{FDC189}} 8.0 & {\cellcolor[HTML]{DD4D04}} 13.5 \\
 & ROUGE-2 & {\cellcolor[HTML]{DD4D04}} 36.4 & {\cellcolor[HTML]{FC8B3A}} 24.8 & {\cellcolor[HTML]{F36E19}} 27.4 & {\cellcolor[HTML]{BB3D02}} 16.7 & {\cellcolor[HTML]{DA4902}} 32.9 & {\cellcolor[HTML]{F67723}} 16.7 & {\cellcolor[HTML]{FDA965}} 9.7 & {\cellcolor[HTML]{FDCA99}} 4.8 & {\cellcolor[HTML]{E5590A}} 26.1 & {\cellcolor[HTML]{FDCFA0}} 6.7 & {\cellcolor[HTML]{F67925}} 10.5 \\
 & ROUGE-L & {\cellcolor[HTML]{DB4B03}} 36.8 & {\cellcolor[HTML]{F57520}} 28.6 & {\cellcolor[HTML]{EA5F0E}} 30.0 & {\cellcolor[HTML]{A53603}} 17.8 & {\cellcolor[HTML]{DB4B03}} 32.4 & {\cellcolor[HTML]{F67925}} 16.5 & {\cellcolor[HTML]{FD984B}} 11.3 & {\cellcolor[HTML]{FDB678}} 6.1 & {\cellcolor[HTML]{862A04}} 37.1 & {\cellcolor[HTML]{FDC189}} 8.0 & {\cellcolor[HTML]{DB4B03}} 13.6 \\
\bottomrule
\end{tabular}

\caption{\textbf{Non Correctness dimensions}: Kendall tau between automatic metrics and human judgement for the dimensions other than Correctness. The top bloc corresponds to coefficients computed when all the human references were available. The second bloc corresponds to coefficients computed given a single human reference. The third bloc corresponds to coefficients computed given no human reference.}
\label{tab:kendall_values_other}
\end{table}

%% file: supp_material/main_table_correctness_kendall_confidence.tex
\begin{table}[h]
\tiny

\begin{tabular}{llcccccccccccc}
\toprule
{} & {} & {WMT} & {Web} & {Asv} & {MUS} & {Pas} & {Fli} & {mSu} & {Rea} & {SumE} & {OpQA} & {OkVQA} & {Avg} \\
\#Ref & Task: & {\cellcolor[HTML]{FFFFFF}} MT & {\cellcolor[HTML]{FFFFFF}} DTG & {\cellcolor[HTML]{FFFFFF}} Sim & {\cellcolor[HTML]{FFFFFF}} Sim & {\cellcolor[HTML]{FFFFFF}} ImCa & {\cellcolor[HTML]{FFFFFF}} ImCa & {\cellcolor[HTML]{FFFFFF}} Sum & {\cellcolor[HTML]{FFFFFF}} Sum & {\cellcolor[HTML]{FFFFFF}} Sum & {\cellcolor[HTML]{FFFFFF}} QA & {\cellcolor[HTML]{FFFFFF}} VQA & {\cellcolor[HTML]{FFFFFF}} All \\
\midrule
\multirow[c]{10}{*}{Max} & ROUGE-1 & {\cellcolor[HTML]{FFF5EB}} 1e-5 & {\cellcolor[HTML]{FFF5EB}} 1e-5 & {\cellcolor[HTML]{FFF5EB}} 1e-5 & {\cellcolor[HTML]{FFF5EB}} 1e-5 & {\cellcolor[HTML]{FFF5EB}} 1e-5 & {\cellcolor[HTML]{FFF5EB}} 1e-5 & {\cellcolor[HTML]{FFF5EB}} 1e-5 & {\cellcolor[HTML]{FFF5EB}} 1e-5 & {\cellcolor[HTML]{FFF5EB}} 1e-5 & {\cellcolor[HTML]{FFF5EB}} 1e-5 & {\cellcolor[HTML]{FFF3E6}} 0.0 & {\cellcolor[HTML]{FFF5EB}} 1e-5 \\
 & ROUGE-2 & {\cellcolor[HTML]{FFF5EB}} 1e-5 & {\cellcolor[HTML]{FFF5EB}} 1e-5 & {\cellcolor[HTML]{FFF5EB}} 1e-5 & {\cellcolor[HTML]{FFF5EB}} 1e-5 & {\cellcolor[HTML]{FFF5EB}} 1e-5 & {\cellcolor[HTML]{FFF5EB}} 1e-5 & {\cellcolor[HTML]{FFF5EB}} 1e-5 & {\cellcolor[HTML]{FFF5EB}} 1e-5 & {\cellcolor[HTML]{FFF5EB}} 1e-5 & {\cellcolor[HTML]{FFF5EB}} 1e-5 & {\cellcolor[HTML]{FDD1A3}} 0.2 & {\cellcolor[HTML]{FFF5EB}} 1e-5 \\
 & ROUGE-L & {\cellcolor[HTML]{FFF5EB}} 1e-5 & {\cellcolor[HTML]{FFF5EB}} 1e-5 & {\cellcolor[HTML]{FFF5EB}} 1e-5 & {\cellcolor[HTML]{FFF5EB}} 1e-5 & {\cellcolor[HTML]{FFF5EB}} 1e-5 & {\cellcolor[HTML]{FFF5EB}} 1e-5 & {\cellcolor[HTML]{FFF5EB}} 1e-5 & {\cellcolor[HTML]{FFF5EB}} 1e-5 & {\cellcolor[HTML]{FFF5EB}} 1e-5 & {\cellcolor[HTML]{FFF5EB}} 1e-5 & {\cellcolor[HTML]{FFF3E6}} 0.0 & {\cellcolor[HTML]{FFF5EB}} 1e-5 \\
\bottomrule
\end{tabular}

\caption{\textbf{Confidence scores - Correctness dimensions}: Kendall tau p-values between automatic metrics and human judgement for Correctness. The top bloc corresponds to coefficients computed when all the human references were available. The second bloc corresponds to coefficients computed given a single human reference. The third bloc corresponds to coefficients computed given no human reference.}
\label{tab:kendall_confidence_correctness}
\end{table}

%% file: supp_material/main_table_other_kendall_confidence.tex
\begin{table}[h]
\tiny

\begin{tabular}{llccccccccccc}
\toprule
{} & {} & {Web} & {Asv} & {Asv} & {MUS} & {mSu} & {SumE} & {SumE} & {SumE} & {OpQA} & {OkVQA} & {OkVQA} \\
\#Ref & Task: & {\cellcolor[HTML]{FFFFFF}} DTG & {\cellcolor[HTML]{FFFFFF}} Sim & {\cellcolor[HTML]{FFFFFF}} Sim & {\cellcolor[HTML]{FFFFFF}} Sim & {\cellcolor[HTML]{FFFFFF}} Sum & {\cellcolor[HTML]{FFFFFF}} Sum & {\cellcolor[HTML]{FFFFFF}} Sum & {\cellcolor[HTML]{FFFFFF}} Sum & {\cellcolor[HTML]{FFFFFF}} QA & {\cellcolor[HTML]{FFFFFF}} VQA & {\cellcolor[HTML]{FFFFFF}} VQA \\
\midrule
\multirow[c]{10}{*}{Max} & ROUGE-1 & {\cellcolor[HTML]{FFF5EB}} 1e-5 & {\cellcolor[HTML]{FFF5EB}} 1e-5 & {\cellcolor[HTML]{FFF5EB}} 1e-5 & {\cellcolor[HTML]{FFF5EB}} 1e-2 & {\cellcolor[HTML]{FFF5EB}} 1e-5 & {\cellcolor[HTML]{FFF5EB}} 1e-5 & {\cellcolor[HTML]{FFF5EB}} 1e-5 & {\cellcolor[HTML]{FFF5EB}} 1e-2 & {\cellcolor[HTML]{FFF5EB}} 1e-5 & {\cellcolor[HTML]{FEE0C1}} 0.1 & {\cellcolor[HTML]{FFF5EB}} 1e-2 \\
 & ROUGE-2 & {\cellcolor[HTML]{FFF5EB}} 1e-5 & {\cellcolor[HTML]{FFF5EB}} 1e-5 & {\cellcolor[HTML]{FFF5EB}} 1e-5 & {\cellcolor[HTML]{FFF4E8}} 0.0 & {\cellcolor[HTML]{FFF5EB}} 1e-5 & {\cellcolor[HTML]{FFF5EB}} 1e-5 & {\cellcolor[HTML]{FFF5EB}} 1e-5 & {\cellcolor[HTML]{FFF4E8}} 0.0 & {\cellcolor[HTML]{FFF5EB}} 1e-5 & {\cellcolor[HTML]{FDC794}} 0.2 & {\cellcolor[HTML]{FFF0E2}} 0.0 \\
 & ROUGE-L & {\cellcolor[HTML]{FFF5EB}} 1e-5 & {\cellcolor[HTML]{FFF5EB}} 1e-5 & {\cellcolor[HTML]{FFF5EB}} 1e-5 & {\cellcolor[HTML]{FFF5EB}} 1e-2 & {\cellcolor[HTML]{FFF5EB}} 1e-5 & {\cellcolor[HTML]{FFF5EB}} 1e-5 & {\cellcolor[HTML]{FFF5EB}} 1e-5 & {\cellcolor[HTML]{FFF5EB}} 1e-2 & {\cellcolor[HTML]{FFF5EB}} 1e-5 & {\cellcolor[HTML]{FEE2C7}} 0.1 & {\cellcolor[HTML]{FFF5EB}} 1e-2 \\
\bottomrule
\end{tabular}

\caption{\textbf{Confidence scores - Non Correctness dimensions}: Kendall tau p-values between automatic metrics and human judgement for the dimensions other than Correctness. The top bloc corresponds to coefficients computed when all the human references were available. The second bloc corresponds to coefficients computed given a single human reference. The third bloc corresponds to coefficients computed given no human reference.}
\label{tab:kendall_confidence_other}
\end{table}

%% file: supp_material/main_table_correctness_pearson_confidence.tex
\begin{table}[h]
\tiny

\begin{tabular}{llcccccccccccc}
\toprule
{} & {} & {WMT} & {Web} & {Asv} & {MUS} & {Pas} & {Fli} & {mSu} & {Rea} & {SumE} & {OpQA} & {OkVQA} & {Avg} \\
\#Ref & Task: & {\cellcolor[HTML]{FFFFFF}} MT & {\cellcolor[HTML]{FFFFFF}} DTG & {\cellcolor[HTML]{FFFFFF}} Sim & {\cellcolor[HTML]{FFFFFF}} Sim & {\cellcolor[HTML]{FFFFFF}} ImCa & {\cellcolor[HTML]{FFFFFF}} ImCa & {\cellcolor[HTML]{FFFFFF}} Sum & {\cellcolor[HTML]{FFFFFF}} Sum & {\cellcolor[HTML]{FFFFFF}} Sum & {\cellcolor[HTML]{FFFFFF}} QA & {\cellcolor[HTML]{FFFFFF}} VQA & {\cellcolor[HTML]{FFFFFF}} All \\
\midrule
\multirow[c]{10}{*}{Max} & ROUGE-1 & {\cellcolor[HTML]{FFF5EB}} 1e-5 & {\cellcolor[HTML]{FFF5EB}} 1e-5 & {\cellcolor[HTML]{FFF5EB}} 1e-5 & {\cellcolor[HTML]{FFF5EB}} 1e-5 & {\cellcolor[HTML]{FFF5EB}} 1e-5 & {\cellcolor[HTML]{FFF5EB}} 1e-5 & {\cellcolor[HTML]{FFF5EB}} 1e-5 & {\cellcolor[HTML]{FFF5EB}} 1e-5 & {\cellcolor[HTML]{FFF5EB}} 1e-5 & {\cellcolor[HTML]{FFF5EB}} 1e-5 & {\cellcolor[HTML]{FFF5EB}} 1e-3 & {\cellcolor[HTML]{FFF5EB}} 1e-5 \\
 & ROUGE-2 & {\cellcolor[HTML]{FFF5EB}} 1e-5 & {\cellcolor[HTML]{FFF5EB}} 1e-5 & {\cellcolor[HTML]{FFF5EB}} 1e-5 & {\cellcolor[HTML]{FFF5EB}} 1e-5 & {\cellcolor[HTML]{FFF5EB}} 1e-5 & {\cellcolor[HTML]{FFF5EB}} 1e-5 & {\cellcolor[HTML]{FFF5EB}} 1e-5 & {\cellcolor[HTML]{FFF5EB}} 1e-5 & {\cellcolor[HTML]{FFF5EB}} 1e-5 & {\cellcolor[HTML]{FFF5EB}} 1e-5 & {\cellcolor[HTML]{FEDEBF}} 0.0 & {\cellcolor[HTML]{FFF5EB}} 1e-5 \\
 & ROUGE-L & {\cellcolor[HTML]{FFF5EB}} 1e-5 & {\cellcolor[HTML]{FFF5EB}} 1e-5 & {\cellcolor[HTML]{FFF5EB}} 1e-5 & {\cellcolor[HTML]{FFF5EB}} 1e-5 & {\cellcolor[HTML]{FFF5EB}} 1e-5 & {\cellcolor[HTML]{FFF5EB}} 1e-5 & {\cellcolor[HTML]{FFF5EB}} 1e-5 & {\cellcolor[HTML]{FFF5EB}} 1e-5 & {\cellcolor[HTML]{FFF5EB}} 1e-5 & {\cellcolor[HTML]{FFF5EB}} 1e-5 & {\cellcolor[HTML]{FFF5EB}} 1e-3 & {\cellcolor[HTML]{FFF5EB}} 1e-5 \\
\bottomrule
\end{tabular}

\caption{\textbf{Confidence scores - Correctness dimensions}: Pearson coefficients p-values between automatic metrics and human judgement for Correctness. The top bloc corresponds to coefficients computed when all the human references were available. The second bloc corresponds to coefficients computed given a single human reference. The third bloc corresponds to coefficients computed given no human reference.}
\label{tab:pearson_confidence_correctness}
\end{table}

%% file: supp_material/main_table_other_pearson_confidence.tex
\begin{table}[h]
\tiny

\begin{tabular}{llccccccccccc}
\toprule
{} & {} & {Web} & {Asv} & {Asv} & {MUS} & {mSu} & {SumE} & {SumE} & {SumE} & {OpQA} & {OkVQA} & {OkVQA} \\
\multirow[c]{2}{*}{\#Ref} & Task: & {\cellcolor[HTML]{FFFFFF}} DTG & {\cellcolor[HTML]{FFFFFF}} Sim & {\cellcolor[HTML]{FFFFFF}} Sim & {\cellcolor[HTML]{FFFFFF}} Sim & {\cellcolor[HTML]{FFFFFF}} Sum & {\cellcolor[HTML]{FFFFFF}} Sum & {\cellcolor[HTML]{FFFFFF}} Sum & {\cellcolor[HTML]{FFFFFF}} Sum & {\cellcolor[HTML]{FFFFFF}} QA & {\cellcolor[HTML]{FFFFFF}} VQA & {\cellcolor[HTML]{FFFFFF}} VQA \\
 & Dim: & {\cellcolor[HTML]{FFFFFF}} Flu & {\cellcolor[HTML]{FFFFFF}} Flu & {\cellcolor[HTML]{FFFFFF}} Sim & {\cellcolor[HTML]{FFFFFF}} Flu & {\cellcolor[HTML]{FFFFFF}} Rel & {\cellcolor[HTML]{FFFFFF}} Rel & {\cellcolor[HTML]{FFFFFF}} Coh & {\cellcolor[HTML]{FFFFFF}} Flu & {\cellcolor[HTML]{FFFFFF}} Obv & {\cellcolor[HTML]{FFFFFF}} Pos & {\cellcolor[HTML]{FFFFFF}} Obv \\
\midrule
\multirow[c]{10}{*}{Max} & ROUGE-1 & {\cellcolor[HTML]{FFF5EB}} 1e-5 & {\cellcolor[HTML]{FFF5EB}} 1e-5 & {\cellcolor[HTML]{FFF5EB}} 1e-5 & {\cellcolor[HTML]{FFF5EB}} 1e-2 & {\cellcolor[HTML]{FFF5EB}} 1e-5 & {\cellcolor[HTML]{FFF5EB}} 1e-5 & {\cellcolor[HTML]{FFF5EB}} 1e-5 & {\cellcolor[HTML]{FFF5EB}} 1e-5 & {\cellcolor[HTML]{FFF5EB}} 1e-5 & {\cellcolor[HTML]{FFF4E9}} 0.0 & {\cellcolor[HTML]{FFF5EB}} 1e-5 \\
 & ROUGE-2 & {\cellcolor[HTML]{FFF5EB}} 1e-5 & {\cellcolor[HTML]{FFF5EB}} 1e-5 & {\cellcolor[HTML]{FFF5EB}} 1e-5 & {\cellcolor[HTML]{FFF5EB}} 1e-2 & {\cellcolor[HTML]{FFF5EB}} 1e-5 & {\cellcolor[HTML]{FFF5EB}} 1e-5 & {\cellcolor[HTML]{FFF5EB}} 1e-5 & {\cellcolor[HTML]{FFF5EB}} 1e-3 & {\cellcolor[HTML]{FFF5EB}} 1e-5 & {\cellcolor[HTML]{FDCB9B}} 0.2 & {\cellcolor[HTML]{FFF5EB}} 1e-2 \\
\bottomrule
\end{tabular}

\caption{\textbf{Confidence scores - Non Correctness dimensions}: Pearson coefficients p-values between automatic metrics and human judgement for the dimensions other than Correctness. The top bloc corresponds to coefficients computed when all the human references were available. The second bloc corresponds to coefficients computed given a single human reference. The third bloc corresponds to coefficients computed given no human reference.}
\label{tab:pearson_confidence_other}
\end{table}